\documentclass{article}

\usepackage{amsmath,amsfonts,bm}

\def\Eqref#1{Equation~\ref{#1}}

\def\1{\bm{1}}

\def\va{{\bm{a}}}

\def\vg{{\bm{g}}}

\def\vl{{\bm{l}}}

\def\vx{{\bm{x}}}

\DeclareMathAlphabet{\mathsfit}{\encodingdefault}{\sfdefault}{m}{sl}
\SetMathAlphabet{\mathsfit}{bold}{\encodingdefault}{\sfdefault}{bx}{n}

\newcommand{\E}{\mathbb{E}}

\DeclareMathOperator*{\argmin}{arg\,min}

\usepackage{amsfonts}
\usepackage{amsmath}
\usepackage{commath}
\usepackage{mathbbol}
\usepackage{cancel}
\usepackage{multirow, pbox}
\usepackage{url}
\usepackage[textwidth=15mm]{todonotes}

\usepackage{times}
\usepackage{graphicx} 
\usepackage[caption=false]{subfig}

\usepackage{algorithm}
\usepackage{algorithmic}

\usepackage{amsthm,amssymb}
\newtheorem{assumption}{Assumption}

\newtheorem{proposition}{Proposition}

\usepackage{microtype}
\usepackage{graphicx}
\usepackage{booktabs} 

\usepackage[accepted, nohyperref]{icml2019}

\icmltitlerunning{Lyapunov-based Safe Policy Optimization for Continuous Control}

\begin{document}

\twocolumn[
\icmltitle{Lyapunov-based Safe Policy Optimization for Continuous Control}



\icmlsetsymbol{equal}{*}

\begin{icmlauthorlist}
\icmlauthor{Yinlam Chow}{dm}
\icmlauthor{Ofir Nachum}{goo}
\icmlauthor{Aleksandra Faust}{goo}
\icmlauthor{Edgar Duenez-Guzman}{dm}
\icmlauthor{Mohammad Ghavamzadeh}{fb}
\end{icmlauthorlist}

\icmlaffiliation{dm}{DeepMind}
\icmlaffiliation{goo}{Google Brain}
\icmlaffiliation{fb}{Facebook AI Research}

\icmlcorrespondingauthor{Yinlam Chow}{yinlamchow@google.com}

\icmlkeywords{Machine Learning, ICML}

\vskip 0.3in
]



\printAffiliationsAndNotice{\icmlEqualContribution} 

\begin{abstract} 
We study continuous action reinforcement learning problems in which it is crucial that the agent interacts with the environment only through {\em safe} policies, i.e.,~policies that do not take the agent to undesirable situations. We formulate these problems as {\em constrained} Markov decision processes (CMDPs) and present safe policy optimization algorithms that are based on a {\em Lyapunov} approach to solve them. Our algorithms can use any standard policy gradient (PG) method, such as deep deterministic policy gradient (DDPG) or proximal policy optimization (PPO), to train a neural network policy, while guaranteeing near-constraint satisfaction for every policy update by projecting either the policy parameter or the action onto the set of feasible solutions induced by the state-dependent linearized Lyapunov constraints. Compared to the existing constrained PG algorithms, ours are more data efficient as they are able to utilize both on-policy and off-policy data. Moreover, our action-projection algorithm often leads to less conservative policy updates and allows for natural integration into an end-to-end PG training pipeline. We evaluate our algorithms and compare them with the state-of-the-art baselines on several simulated (MuJoCo) tasks, as well as a real-world indoor robot navigation problem, demonstrating their effectiveness in terms of balancing performance and constraint satisfaction. Videos of the experiments can be found in the following link: \url{https://drive.google.com/file/d/1pzuzFqWIE710bE2U6DmS59AfRzqK2Kek/view?usp=sharing}

\end{abstract}
\vspace{-0.2in}

\vspace{-0.05in}
\section{Introduction}\label{sec:intro}
\vspace{-0.05in}

The field of reinforcement learning (RL) has witnessed tremendous success in many high-dimensional control problems, including video games~\citep{Mnih15HL}, board games~\citep{Silver16MG}, 
robot locomotion~\citep{Lillicrap16CC}, manipulation~\citep{Levine16EE,qtopt}, navigation~\citep{faust2018prm}, and obstacle avoidance~\citep{autorl}. In standard RL, the ultimate goal is to optimize the expected sum of rewards/costs, and the agent is free to explore any behavior as long as it leads to performance improvement. Although this freedom might be acceptable in many problems, including those involving simulated environments, and could expedite learning a good policy, it might be harmful in many other problems and could cause damage to the agent (robot) or to the environment (plant or the people working nearby). In such domains, it is absolutely crucial that while the agent (RL algorithm) optimizes its long-term performance, it also maintains safe policies both during training and at convergence. 

A natural way to incorporate safety is via constraints. A standard model for RL with constraints is constrained Markov decision process (CMDP)~\citep{altman1999constrained}, where in addition to its standard objective, the agent must satisfy constraints on expectations of auxiliary costs. Although optimal policies for finite CMDPs with known models can be obtained by linear programming~\citep{altman1999constrained}, there are not many results for solving CMDPs when the model is unknown or the state and/or action spaces are large or infinite. A common approach to solve CMDPs is to use the Lagrangian method~\citep{altman1998constrained,geibel2005risk} that augments the original objective function with a penalty on constraint violation and computes the saddle-point of the constrained policy optimization via primal-dual methods~\citep{chow2017risk}. Although safety is ensured when the policy converges asymptotically, a major drawback of this approach is that it makes no guarantee with regards to the safety of the policies generated during training. 

A few algorithms have been recently proposed to solve CMDPs at scale, while remaining safety during training. One such algorithm is {\em constrained policy optimization} (CPO)~\citep{achiam2017constrained}. CPO extends the trust-region policy optimization (TRPO) algorithm~\citep{Schulman15TR} to handle the constraints in a principled way and has shown promising empirical results in terms scalability, performance, and constraint satisfaction, both during training and after convergence. Another class of algorithms of this sort is by~\citet{chow2018lyapunov}. These algorithms use the notion of Lyapunov functions that have a long history in control theory to analyze the stability of dynamical systems~\citep{khalil1996noninear}. Lyapunov functions have been used in RL to guarantee closed-loop stability of the agent~\citep{perkins2002lyapunov,faust-acta-14}. They also have been used to guarantee that a model-based RL agent can be brought back to a ``region of attraction'' during exploration~\citep{berkenkamp2017safe}.~\citet{chow2018lyapunov} 
use the theoretical properties of the Lyapunov functions and propose safe approximate policy and value iteration algorithms. They prove theories for their algorithms, when the CMDP is finite and known, and empirically evaluate them when it is large and/or unknown. However, since their algorithms are value-function-based, applying them to continuous action problems is not straightforward, and was left as a future work. 

In this paper, we build on the problem formulation and theoretical findings of the Lyapunov-based approach to solve CMDPs, and extend it to tackle continuous action problems that play an important role in control theory and robotics. We propose Lyapunov-based safe RL algorithms that can handle problems with large or infinite action spaces, and return safe policies both during training and at convergence. 
To do so, there are two major difficulties which we resolve: 
{\bf 1)} the policy update becomes an optimization problem over the large or continuous action space (similar to standard MDPs with large actions), and {\bf 2)} the policy update is a constrained optimization problem in which the (Lyapunov) constraints involve integration over the action space, and thus, it is often impossible to have them in closed-form. Since the number of Lyapunov constraints is equal to the number of states, the situation is even more challenging when the problem has a large or an infinite state space. To address the first difficulty, we switch from value-function-based to policy gradient (PG) and actor-critic algorithms. To address the second difficulty, we propose two approaches to solve our constrained policy optimization problem (a problem with infinite constraints, each involving an integral over the continuous action space) that can work with any standard on-policy (e.g.,~proximal policy optimization (PPO)~\citealt{schulman2017proximal}) and off-policy (e.g.,~deep deterministic policy gradient (DDPG)~\citealt{lillicrap2015continuous}) PG algorithm. Our first approach, which we call {\em policy parameter projection} or {\em $\theta$-projection}, is a constrained optimization method that combines PG with a projection of the policy parameters onto the set of feasible solutions induced by the Lyapunov constraints. Our second approach, which we call {\em action projection} or {\em $a$-projection}, uses the concept of a {\em safety layer} introduced by~\citet{dalal2018safe} to handle simple single-step constraints, extends this concept to general trajectory-based constraints, solves the constrained policy optimization problem in closed-form using Lyapunov functions, and integrates this closed-form into the policy network via safety-layer augmentation. Since both approaches guarantee safety at every policy update, they manage to maintain safety throughout training (ignoring errors resulting from function approximation), ensuring that all intermediate policies are safe to be deployed. To prevent constraint violations due to function approximation and modeling errors, similar to CPO, we offer a safeguard policy update rule that decreases constraint cost and ensures near-constraint satisfaction.

Our proposed algorithms have two main advantages over CPO. First, since CPO is closely connected to TRPO, it can only be trivially combined with PG algorithms that are regularized with relative entropy, such as PPO. This restricts CPO to on-policy PG algorithms. On the contrary, our algorithms can work with any on-policy (e.g.,~PPO) and off-policy (e.g.,~DDPG) PG algorithm. Having an off-policy implementation is beneficial, since off-policy algorithms are potentially more data-efficient, as they can use the data from the replay buffer. Second, while CPO is not a \emph{back-propagatable} algorithm, due to the backtracking line-search procedure and the conjugate gradient iterations for computing natural gradient in TRPO, our algorithms can be trained \emph{end-to-end}, which is crucial for scalable and efficient implementation~\citep{hafner2017tensorflow}. In fact, we show in Section~\ref{sec:theta_proj} that CPO (minus the line search) can be viewed as a special case of the on-policy version (PPO version) of our $\theta$-projection algorithm, corresponding to a specific approximation of the constraints. 

We evaluate our algorithms and compare them with CPO and the Lagrangian method on several continuous control  (MuJoCo) tasks and a real-world robot navigation problem, in which the robot must satisfy certain constraints, while minimizing its expected cumulative cost. Results show that our algorithms outperform the baselines in terms of balancing the performance and constraint satisfaction (during training), and generalize better to new and more complex environments, including transfer to a real Fetch robot.

\vspace{-0.1in}
\section{Preliminaries}\label{sec:prelim}
\vspace{-0.05in}

We consider the RL problem in which the agent's interaction with the environment is modeled as a Markov decision process (MDP). A MDP is a tuple $(\mathcal X,\mathcal A,\gamma, c,P,x_0)$, where $\mathcal X$ and $\mathcal A$ are the state and action spaces; $\gamma\in[0,1)$ is a discounting factor; $c(x,a)\in[0,C_{\max}]$ is the immediate cost function; $P(\cdot|x,a)$ is the transition probability distribution; and $x_0\in \mathcal X$ is the initial state. Although we consider deterministic initial state and cost function, our results can be easily generalized to random initial states and costs. We model the RL problems in which there are constraints on the cumulative cost using CMDPs. The CMDP model extends MDP by introducing additional costs and the associated constraints, and is defined by $(\mathcal X,\mathcal A,\gamma,c,P,x_0,d,d_0)$, where the first six components are the same as in the unconstrained MDP; $d(x)\in[0,D_{\max}]$ is the (state-dependent) immediate constraint cost; and $d_0 \in \mathbb{R}_{\geq 0}$ is an upper-bound on the expected cumulative constraint cost.

To formalize the optimization problem associated with CMDPs, let $\Delta$ be the set of Markovian stationary policies, i.e.,~$\Delta=\big\{\pi:\mathcal X\times \mathcal A\rightarrow [0,1],\;\sum_{a}\pi(a|x)=1\big\}$. At each state $x\in\mathcal X$, we define the generic Bellman operator w.r.t.~a policy $\pi\in\Delta$ and a cost function $h$ as $T_{\pi,h}[V](x)=\sum_{a}\pi(a|x)\![h(x,a)\!+\!\gamma\sum_{x'\in\mathcal X}\!P(x'|x,a)V(x')]$. Given a policy $\pi\in \Delta$, we define the expected cumulative cost and the safety constraint function (expected cumulative constraint cost) as $\mathcal C_\pi(x_0):=\mathbb E\big[\sum_{t=0}^{\infty}\gamma^t c(x_t, a_t)\mid \pi,x_0\big]$ and $\mathcal D_\pi(x_0):=\mathbb E\big[\sum_{t=0}^{\infty}\gamma^td(x_t)\mid \pi, x_0\big]$. The {\em safety constraint} is then defined as $\mathcal D_\pi(x_0)\leq d_0$.
The goal in CMDPs is to solve the constrained optimization problem 
%
\begin{equation}
\label{eq:CMDP}
\pi^*\in\min_{\pi \in \Delta} \,\left\{\mathcal C_\pi(x_0): \mathcal D_\pi(x_0)\leq d_0\right\}. 
\end{equation}
%
It has been shown that if the feasibility set is non-empty, then there exists an optimal policy in the class of stationary Markovian policies $\Delta$~\citep[Theorem~3.1]{altman1999constrained}. 

\vspace{-0.1in}
\subsection{Policy Gradient Algorithms}
\vspace{-0.05in}

Policy gradient (PG) algorithms optimize a policy by computing a sample estimate of the gradient of the expected cumulative cost induced by the policy, and then updating the policy parameter in the gradient direction. In general, stochastic policies that give a probability distribution over actions are parameterized by a $\kappa$-dimensional vector $\theta$, so the space of policies can be written as $\big\{\pi_\theta,\;\theta\in\Theta\subset\mathbb R^{\kappa}\big\}$. Since in this setting a policy $\pi$ is uniquely defined by its parameter $\theta$, policy-dependent functions can be written as a function of $\theta$ or $\pi$ interchangeably. 

Deep deterministic policy gradient (DDPG)~\citep{lillicrap2015continuous} and proximal policy optimization (PPO)~\citep{schulman2017proximal} are two PG algorithms that have recently gained popularity in solving continuous control problems. DDPG is an off-policy Q-learning style algorithm that jointly trains a deterministic policy $\pi_\theta(x)$ and a Q-value approximator $Q(x,a;\phi)$. The Q-value approximator is trained to fit the true Q-value function 
and the deterministic policy is trained to optimize $Q(x,\pi_\theta(x);\phi)$ via chain-rule. The PPO algorithm we use is a penalty form of the trust region policy optimization (TRPO) algorithm~\citep{Schulman15TR} with an adaptive rule to tune the $D_{KL}$ penalty weight $\beta_k$. PPO trains a policy $\pi_\theta(x)$ by optimizing a loss function that consists of the standard policy gradient objective and a penalty on the KL-divergence between the current $\theta$ and previous $\theta'$ policies, i.e.,~$\overline D_{\mathrm{KL}}(\theta, \theta')=\E[\sum_t\gamma^tD_{\mathrm{KL}}(\pi_{\theta'}(\cdot|x_t) || \pi_\theta(\cdot|x_t))| \pi_{\theta'}, x_0]$.

\vspace{-0.1in}
\subsection{Lagrangian Method}
\vspace{-0.05in}

The Lagrangian method is a straightforward way to address the constraint $\mathcal D_{\pi_\theta}(x_0)\leq d_0$ in CMDPs. In this approach, we add the constraint costs $d(x)$ to the task costs $c(x,a)$ and transform the constrained optimization problem to a penalty form, i.e.,~$\min_{\theta\in\Theta} \max_{\lambda\ge0} \E\left[\sum_{t=0}^\infty c(x_t,a_t)+\lambda d(x_t)| \pi_\theta, x_0\right] - \lambda d_0$. 
We then jointly optimizes $\theta$ and $\lambda$ to find a saddle-point of the penalized objective. The optimization of $\theta$ may be performed by any PG algorithm, such as DDPG or PPO, on the augmented cost $c(x,a) + \lambda d(x)$, while $\lambda$ is optimized by stochastic gradient descent. As described in Section~\ref{sec:intro}, although the Lagrangian approach is easy to implement (see Appendix~\ref{appendix:lag} for the details), in practice, it often violates the constraints during training. While at each step during training, the objective encourages finding a safe solution, the current value of $\lambda$ may lead to an unsafe policy. This is why the Lagrangian method may not be suitable for solving problems in which safety is crucial during training. 

\vspace{-0.1in}
\subsection{Lyapunov Functions}
\vspace{-0.05in}

Since in this paper, we extend the Lyapunov-based approach to CMDPs to PG algorithms, we end this section by introducing some terms and notations from~\citet{chow2018lyapunov} that are important in developing our safe PG algorithms. We refer the reader to Appendix~\ref{sec:lyapunov_cmdp} for more details. 

We define a set of Lyapunov functions w.r.t.~initial state $x_0\in\mathcal X$ and constraint threshold $d_0$ as $\mathcal L_{\pi_B}(x_0,d_0)=\{L:\mathcal X\rightarrow\mathbb R_{\geq 0}\mid T_{\pi_{B},d}[L](x)\leq L(x),\; \forall x\in\mathcal X,\; L(x_0)\leq d_0 \}$, where $\pi_B$ is a feasible policy of~\eqref{eq:CMDP}, i.e.,~$\mathcal D_{\pi_B}(x_0)\leq d_0$. We refer to the constraints in this feasibility set as the {\em Lyapunov constraints}. For any arbitrary Lyapunov function $L\in\mathcal L_{\pi_{B}}(x_0,d_0)$, we denote by $\mathcal F_{L}=\left\{\pi\in\Delta:T_{\pi,d}[L](x)\leq L(x),\;\forall x\in\mathcal X\right\}$, the set of $L$-induced Markov stationary policies. The contraction property of $T_{\pi,d}$, together with $ L(x_0)\leq d_0$, imply that any $L$-induced policy is a feasible policy of~\eqref{eq:CMDP}. However, $\mathcal F_{L}(x)$ does not always contain an optimal solution of~\eqref{eq:CMDP}, and thus, it is necessary to design a Lyapunov function that provides this guarantee. In other words, the main goal of the Lyapunov approach is to construct a Lyapunov function $L\in \mathcal L_{\pi_B}(x_0,d_0)$, such that $\mathcal F_{L}$ contains an optimal policy $\pi^*$, i.e., $L(x)\geq T_{\pi^*,d}[L](x)$.~\citet{chow2018lyapunov} show in their Theorem~1 that without loss of optimality, the Lyapunov function that satisfies the above criterion can be expressed as $L_{\pi_B,\epsilon}(x):=\mathbb E\big[\sum_{t=0}^\infty \gamma^t\big(d(x_t)+\epsilon(x_t)\big)\mid \pi_{B},x\big]$, in which $\epsilon(x)\geq 0$ is a specific immediate {\em auxiliary constraint cost} that keeps track of the maximum \emph{constraint budget} available for policy improvement (from $\pi_B$ to $\pi^*$). 
They propose ways to construct such $\epsilon$, as well as an auxiliary constraint cost surrogate $\widetilde{\epsilon}$, which is a tight upper-bound on $\epsilon$ and can be computed more efficiently. 
They use this construction to propose their safe (approximate) policy and value iteration algorithms, in which the goal is to solve the following LP problem~\citep[Eq.~6]{chow2018lyapunov} at each policy improvement step:

\vspace{-0.25in}
\begin{small}
\begin{align}
\label{eq:main-opt-prob}
&\pi_+(\cdot|x)=\argmin_{\pi\in\Delta} \; \int_{a\in\mathcal A}Q_{V_{\pi_B}}(x, a)\pi(a|x), \;\; \text{subject to} \\
&\int_{a\in\mathcal A}\; Q_{L_{\pi_B}}(x, a)\;\big(\pi(a|x)-\pi_B(a|x)\big)\leq \widetilde\epsilon(x),\;\;\forall x\in\mathcal X, \nonumber
\end{align}
\end{small}
\vspace{-0.225in}

where $V_{\pi_B}(x)=T_{\pi_{B},c}[V_{\pi_B}](x)$ and $Q_{V_{\pi_B}}(x,a)=c(x,a)+\gamma\sum_{x'}P(x'|x,a)V_{\pi_B}(x')$ are the value function and state-action value function (w.r.t.~the cost function $c$), and $Q_{L_{\pi_B}}(x,a)=d(x)+\widetilde\epsilon(x)+\gamma\sum_{x'}P(x'|x,a)L_{\pi_B,\widetilde\epsilon}(x')$ is the Lyapunov function. Note that in an iterative policy optimization method, such as those we will present in this paper, the feasible policy $\pi_B$ can be set to the policy at the previous iteration. 

In~\eqref{eq:main-opt-prob}, there are as many constraints as the number of states and each constraint involves an integral over the entire action space $\mathcal A$. When the state space is large or continuous, even if the integral in the constraint has a closed-form (e.g.,~when the number of actions is finite), solving LP~\eqref{eq:main-opt-prob} becomes numerically intractable. Since~\citet{chow2018lyapunov} assume that the number of actions is finite, they focus on value-function-based RL algorithms and address the large state issue by \emph{policy distillation}. However, in this paper, we are interested in problems with large action spaces. In our case, solving~\eqref{eq:main-opt-prob} will be even more challenging. To address this issue, in the next section, we first switch from value-function-based algorithms to PG algorithms, then propose an optimization problem with Lyapunov constraints, analogous to~\eqref{eq:main-opt-prob}, that is suitable for the PG setting, and finally present two methods to solve our proposed optimization problem efficiently.



\vspace{-0.05in}
\section{Safe Lyapunov-based Policy Gradient}\label{sec:algorthms}


We now present our approach to solve CMDPs in a way that guarantees safety both at convergence and during training.
Similar to~\citet{chow2018lyapunov}, our Lyapunov-based safe PG algorithms solve a constrained optimization problem analogous to~\eqref{eq:main-opt-prob}. In particular, our algorithms consist of two components, a baseline PG algorithm, such as DDPG or PPO, and an effective method to solve the general Lyapunov-based policy optimization problem (the analogous to~\eqref{eq:main-opt-prob})

\vspace{-0.25in}
\begin{small}
\begin{align}
\label{eq:main-opt-prob_pg}
&\theta_+=\argmin_{\theta\in\Theta} \; \mathcal C_{\pi_\theta}(x_0), \quad \text{subject to} \\
&\int_{a\in\mathcal A}\big(\pi_\theta(a|x)-\pi_B(a|x)\big)\; Q_{L_{\pi_B}}(x, a)\;da\leq \widetilde\epsilon(x),\;\;\forall x\in\mathcal X. \nonumber
\end{align}
\end{small}
\vspace{-0.25in}

In the next two sections, we present two approaches to solve~\eqref{eq:main-opt-prob_pg} efficiently. We call these approaches {\bf 1)} $\theta$-projection, a constrained optimization method that combines PG with projecting the policy parameter $\theta$ onto the set of feasible solutions induced by the Lyapunov constraints, and {\bf 2)} $a$-projection, in which we embed the Lyapunov constraints into the policy network via a safety layer. 


\vspace{-0.05in}
\subsection{The $\theta$-projection Approach}\label{sec:theta_proj}


In this section, we show how a safe Lyapunov-based PG algorithm can be derived using the $\theta$-projection approach. This machinery is based on the \emph{minorization-maximization} technique in conservative PG~\citep{kakade2002approximately} and Taylor series expansion, and it can be applied to both on-policy and off-policy algorithms. 
Following Theorem 4.1 in~\citet{kakade2002approximately}, we first have the following bound for the cumulative cost: $- \beta \overline D_{\mathrm{KL}}(\theta, \theta_B) \leq \mathcal C_{\pi_\theta}(x_0)- \mathcal C_{\pi_{\theta_B}}(x_0) - \mathbb{E}_{x\sim\mu_{\theta_B,x_0},a\sim \pi_\theta}\big[Q_{V_{\theta_B}}(x,a)-V_{\theta_B}(x)\big]\leq \beta \overline D_{\mathrm{KL}}(\theta, \theta_B)$, where $\mu_{\theta_B,x_0}$ is the $\gamma$-visiting distribution of $\pi_{\theta_B}$ starting at the initial state $x_0$, and $\beta$ is the weight for the entropy-based regularization.\footnote{Theorem~1 in~\citet{Schulman15TR} provides a recipe for computing $\beta$ such that the minorization-maximization inequality holds. But in practice, $\beta$ is treated as a tunable hyper-parameter for entropy-based regularization.} Using the above result, we denote by 

\vspace{-0.25in}
\[
\begin{split}
\mathcal C'_{\pi_\theta}(x_0;\pi_{\theta_B})=& \mathcal C_{\pi_{\theta_B}}(x_0) + \beta \overline D_{\mathrm{KL}}(\theta, \theta_B) +\\ 
&\mathbb{E}_{x\sim\mu_{\theta_B,x_0},a\sim \pi_\theta}\big[Q_{V_{\theta_B}}(x,a)-V_{\theta_B}(x)\big]
\end{split}
\]
\vspace{-0.2in}

the surrogate cumulative cost. It has been shown in Eq.~10 of~\citet{Schulman15TR} that replacing the objective function $\mathcal C_{\pi_\theta}(x_0)$ with its surrogate $\mathcal C'_{\pi_\theta}(x_0;\pi_{\theta_B})$ in solving~\eqref{eq:main-opt-prob_pg} will still lead to policy improvement.
In order to effectively compute the improved policy parameter $\theta_+$, one further approximates the function $\mathcal C'_{\pi_\theta}(x_0;\pi_{\theta_B})$ with its Taylor series expansion (around $\theta_B$). In particular, the term $\mathbb{E}_{x\sim\mu_{\theta_B,x_0},a\sim \pi_\theta}\big[Q_{V_{\theta_B}}(x,a)-V_{\theta_B}(x)\big]$ is approximated up to its first order, 
and the term $\overline D_{\mathrm{KL}}(\theta, \theta_B)$ is approximated up to its second order. 
Altogether this allows us to replace the objective function in~\eqref{eq:main-opt-prob_pg} with the following surrogate:

\vspace{-0.25in}
\[
\begin{split}
&\langle(\theta-\theta_B),\nabla_\theta\mathbb E_{x\sim \mu_{\theta_B,x_0},a\sim \pi_\theta} \big[Q_{V_{\theta_B}}(x,a)\big]\rangle \\
&\quad+ \frac{\beta}{2}\langle(\theta-\theta_B),\nabla^2_\theta \overline D_{\mathrm{KL}}(\theta, \theta_B)\mid_{\theta=\theta_B}(\theta-\theta_B)\rangle.
\end{split}
\]
\vspace{-0.2in}

Similarly, regarding the constraints in~\eqref{eq:main-opt-prob_pg}, we can use the Taylor series expansion (around $\theta_B$) to approximate the LHS of the Lyapunov constraints as 

\vspace{-0.25in}
\[
\begin{split}
&\int_{a\in\mathcal A}\big(\pi_\theta(a|x)-\pi_B(a|x)\big)\; Q_L(x,a)\;da\approx\\
&\quad\quad\big\langle(\theta-\theta_B), \nabla_\theta\mathbb E_{a\sim \pi_\theta} \big[Q_{L_{\theta_B}}(x,a)\big]\mid_{\theta=\theta_B}\big\rangle.
\end{split}
\]
\vspace{-0.2in}

Using the above approximations, at each iteration, our safe PG algorithm updates the policy by solving the following constrained optimization problem with \emph{semi-infinite dimensional} Lyapunov constraints:

\vspace{-0.25in}
\begin{small}
\begin{align}
\label{eq:opt-prob2}
&\theta_+\in\argmin_{\theta \in\Theta}\; \Big\langle(\theta-\theta_B),\nabla_\theta\mathbb E_{x\sim \mu_{\theta_B,x_0},a\sim \pi_\theta} \big[Q_{V_{\theta_B}}(x,a)\big]\Big\rangle \nonumber \\
&\quad\quad+\frac{\beta}{2}\Big\langle(\theta-\theta_B),\nabla^2_\theta \overline D_{\mathrm{KL}}(\theta, \theta_B)\mid_{\theta=\theta_B}(\theta-\theta_B)\Big\rangle, \\
&\text{s.t. } \Big\langle\!(\theta-\theta_B), \nabla_\theta\mathbb E_{a\sim \pi_\theta} \big[Q_{L_{\theta_B}}(x,a)\big]\mid_{\theta=\theta_B}\!\!\Big\rangle\leq \widetilde\epsilon(x),\;\forall x\in\mathcal X. \nonumber
\end{align}
\end{small}
\vspace{-0.25in}

It can be seen that if the errors resulted from the neural network parameterizations of $Q_{V_{\theta_B}}$ and $Q_{L_{\theta_B}}$, and the Taylor series expansions are small, then an algorithm that updates the policy parameter by solving~\eqref{eq:opt-prob2} can ensure safety during training. However, the presence of infinite-dimensional Lyapunov constraints makes solving~\eqref{eq:opt-prob2} numerically intractable. A solution to this is to write the Lyapunov constraints in~\eqref{eq:opt-prob2} (without loss of optimality) as
$
\max_{x\in\mathcal X}\langle(\theta-\theta_B), \nabla_\theta\mathbb E_{a\sim \pi_\theta} [Q_{L_{\theta_B}}(x,a)]\mid_{\theta=\theta_B}\rangle- \widetilde\epsilon(x)\le 0.
$
Since the above $\max $-operator is non-differentiable, this may still lead to numerical instability in gradient descent algorithms. Similar to the surrogate constraint used in TRPO (to transform the $\max D_{\text{KL}}$ constraint to an average $\overline D_{\text{KL}}$ constraint, see Eq.~12 in~\citealt{Schulman15TR}), a more numerically stable way is to \emph{approximate} the Lyapunov constraint using the following average constraint surrogate:

\vspace{-0.25in}
\begin{small}
\begin{equation}
\label{eq:avg_constraint_threshold}
\Big\langle\!(\theta-\theta_B),\frac{1}{M}\!\sum_{i=1}^M\nabla_\theta\mathbb E_{a\sim \pi_\theta} \big[Q_{L_{\theta_B}}(x_i,a)\big]\mid_{\theta=\theta_B}\!\!\!\Big\rangle\!\leq\! \frac{1}{M}\!\sum_{i=1}^M\widetilde\epsilon(x_i),
\end{equation}
\end{small}
\vspace{-0.25in}

where $M$ is the number of on-policy sample trajectories of $\pi_{\theta_B}$. In practice, when the auxiliary constraint surrogate is chosen as  $\widetilde\epsilon=(1-\gamma)\big(d_0-\mathcal D_{\pi_{\theta_B}}(x_0)\big)$  (see Appendix \ref{sec:lyapunov_cmdp} for the justification of this choice), the gradient term in~\eqref{eq:avg_constraint_threshold} can be simplified as $\nabla_\theta\mathbb E_{a\sim \pi_\theta} \big[Q_{L_{\theta_B}}(x_i,a)\big]=\nabla_\theta\int_{a}\pi_\theta(a|x)\,\nabla_\theta \log\pi_\theta(a|x) \,Q_{W_{\theta_B}}(x_i,a)da$, where $W_{\theta_B}(x)=T_{\pi_{B},d}[W_{\theta_B}](x)$ and $Q_{W_{\theta_B}}(x,a)=d(x)+\gamma\sum_{x'}P(x'|x,a)W_{\theta_B}(x')$ are the constraint value function and constraint state-action value function, respectively. Combining with the fact that $\widetilde\epsilon$ is state independent, the above arguments further imply that the average constraint surrogate in~\eqref{eq:avg_constraint_threshold} can be approximated by the inequality $\mathcal{D}_{\pi_{\theta_B}}(x_0)+\frac{1}{1-\gamma}\langle(\theta-\theta_B),\frac{1}{M}\!\sum_{i=1}^M\!\nabla_\theta\mathbb E_{a\sim\pi_\theta}\!\big[Q_{W_{\theta_B}}(x_i,a)\big]\!\!\mid_{\theta=\theta_B}\rangle\!\leq\! d_0$, which is equivalent to the constraint used in CPO (see Sec.~6.1 in~\citealt{achiam2017constrained}). This shows a clear connection between CPO (minus the line search) and our Lyapunov-based PG with $\theta$-projection. Algorithm~\ref{alg:cpo} in Appendix~\ref{appendix:alg} contains the pseudo-codes of our safe Lyapunov-based PG algorithms with $\theta$-projection. We refer to the DDPG and PPO versions of this algorithm as SDDPG and SPPO. 


\vspace{-0.05in}
\subsection{The $a$-projection Approach}\label{sec:a_proj}

Note that the main characteristic of the Lyapunov approach is to break down a trajectory-based constraint into a sequence of single-step \emph{state dependent} constraints. However, when the state space $\mathcal X$ is infinite, the feasibility set is characterized by infinite dimensional constraints, and thus, it is actually counter-intuitive to directly enforce these Lyapunov constraints (as opposed to the original trajectory-based constraint) into the policy update optimization. To address this issue, we leverage the idea of a \emph{safety layer} from~\citet{dalal2018safe}, that was applied to simple single-step constraints, and propose a novel approach to embed the set of Lyapunov constraints into the policy network. This way, we reformulate the CMDP problem~\eqref{eq:CMDP} as an unconstrained optimization problem and optimize its policy parameter $\theta$ (of the augmented network) using any standard unconstrained PG algorithm. At every given state, the unconstrained action is first computed and then passed through the safety layer, where a feasible action mapping is constructed by projecting the unconstrained actions onto the feasibility set w.r.t.~the corresponding Lyapunov constraint. Therefore, safety during training w.r.t.~the CMDP problem can be guaranteed by this \emph{constraint projection approach}. 

For simplicity, we only describe how the action mapping (to the set of Lyapunov constraints) works for deterministic policies. Using identical machinery, this procedure can be extended to guarantee safety for stochastic policies. Recall from the policy improvement problem in~\eqref{eq:main-opt-prob_pg} that the Lyapunov constraint is imposed at every state $x\in\mathcal X$. 
Given a baseline feasible policy $\pi_B=\pi_{\theta_B}$, for any arbitrary policy parameter $\theta\in\Theta$, we denote by $\Xi(\pi_B,\theta)=\big\{\theta'\in\Theta:Q_{L_{\pi_B}}(x, \pi_{\theta'}(x))-Q_{L_{\pi_B}}(x, \pi_B(x))\leq \widetilde\epsilon(x),\,\forall x\in\mathcal X\big\}$, the \emph{projection} of $\theta$ onto the feasibility set induced by the Lyapunov constraints. One way to construct a feasible policy $\pi_{\Xi(\pi_B,\theta)}$ from a parameter $\theta$ is to solve the following $\ell_2$-projection problem at each state $x\in\mathcal X$:
\begin{align}
\label{opt:safety_layer_ori}
\pi_{\Xi(\pi_B,\theta)}(x) &\in \argmin_{a} \;\frac{1}{2}\|a-\pi_{\theta}(x)\|^2, \\
&\text{s.t. } \; Q_{L_{\pi_B}}(x, a)-Q_{L_{\pi_B}}(x, \pi_B(x))\leq \widetilde\epsilon(x). \nonumber
\end{align}
We refer to this operation as the \emph{Lyapunov safety layer}. Intuitively, this projection perturbs the unconstrained action as little as possible in the Euclidean norm in order to satisfy the Lyapunov constraints. Since this projection guarantees safety (in the Lyapunov sense), if we have access to a closed form of the projection, we may insert it into the policy parameterization and simply solve an 
unconstrained policy optimization problem, i.e.,~$\theta_+\in\argmin_{\theta\in\Theta} \; \mathcal C_{\pi_{\Xi(\pi_B,\theta)}}(x_0)$, using any standard PG algorithm.

To simplify the projection~\eqref{opt:safety_layer_ori}, we can approximate the LHS of the Lyapunov constraint with its first-order Taylor series (w.r.t.~action $a=\pi_B(x)$). Thus, at any given state $x\in\mathcal X$, the safety layer solves the following projection problem:
\begin{align}
\label{opt:safety_layer}
\pi_{\Xi(\pi_B,\theta)}(x) &\in \argmin_{a} \frac{1}{2}\|a-\pi_{\theta,\text{unc}}(x)\|^2, \\
&\text{s.t. } \quad \big(a-\pi_B(x)\big)^\top g_{L_{\pi_B}}(x)\leq\widetilde\epsilon(x), \nonumber
\end{align}
where $g_{L_{\pi_B}}(x):=\nabla_a Q_{L_{\pi_B}}(x, a)\mid_{a=\pi_B(x)}$ is the action-gradient of the state-action Lyapunov function induced by the baseline action $a=\pi_B(x)$. 

Similar to the analysis of Section~\ref{sec:theta_proj}, if the auxiliary cost $\widetilde\epsilon$ is state independent, one can readily find $ g_{L_{\pi_B}}(x)$ by computing the gradient of the constraint action-value function $\nabla_a Q_{W_{\theta_B}}(x, a)\mid_{a=\pi_B(x)}$. Note that the objective function in~\eqref{opt:safety_layer} is positive-definite and quadratic, and the constraint approximation is linear. Therefore, the solution of this (convex) projection problem can be effectively computed by an in-graph QP-solver, such as OPT-Net~\citep{amos2017optnet}. Combined with the  above projection procedure, this further implies that the CMDP problem can be effectively solved using an \emph{end-to-end} PG training pipeline (such as DDPG or PPO). Furthermore, when the CMDP has a single constraint (and thus a single Lyapunov constraint), the policy $\pi_{\Xi(\pi_B,\theta)}(x)$ has the following analytical solution.
\begin{proposition}
At any given state $x\in\mathcal X$, the solution to the optimization problem~\eqref{opt:safety_layer} has the form 
$\pi_{\Xi(\pi_B,\theta)}(x)=\pi_{\theta}(x)+\lambda^*(x) \cdot g_{L_{\pi_B}}(x)$, where

\vspace{-0.15in}
\begin{small}
\begin{equation*}
\lambda^*(x)=\left(\frac{g_{L_{\pi_B}}(x)^\top\big(\pi_{\theta}(x)-\pi_B(x)\big)-\widetilde\epsilon(x)}{g_{L_{\pi_B}}(x)^\top g_{L_{\pi_B}}(x)}\right)_+.
\end{equation*}
\end{small}
\vspace{-0.25in}

\end{proposition}
The closed-form solution is essentially a linear projection of the
unconstrained action $\pi_{\theta}(x)$ to the Lyapunov-safe hyperplane characterized with slope
$ g_{L_{\pi_B}}(x)$ and intercept $\widetilde\epsilon(x)=(1-\gamma) (d_0-\mathcal D_{\pi_B}(x_0))$.
Extending this closed-form solution to handle multiple constraints is possible, if there is at most one constraint active at a time (see Proposition~1 in~\citealt{dalal2018safe} for a similar extension). 

Without loss of generality, this projection step can also be extended to handle actions generated by stochastic policies with bounded first and second order moments~\citep{yu2009general}. For example when the policy is parameterized with a Gaussian distribution, then one needs to project both the mean and standard-deviation vector onto the Lyapunov-safe hyperplane, in order to obtain a feasible action probability. Algorithm~\ref{alg:safety_layer} in Appendix~\ref{appendix:alg} contains the pseudo-code of our safe Lyapunov-based PG algorithms with $a$-projection. We refer to the DDPG and PPO versions of this algorithm as SDDPG-modular and SPPO-modular, respectively.


\vspace{-0.05in}
\section{Experiments on MuJoCo Benchmarks}\label{sec:experiements}
\vspace{-0.05in}

\begin{figure}[tb]
\centering
  \begin{tabular}{cc}
      \subfloat[\scriptsize HalfCheetah-Safe, Return]{\label{fig:hc-ret}\includegraphics[trim=3mm 3mm 7mm 3mm,clip,width=0.22\textwidth, height=2.2cm,keepaspectratio=false]{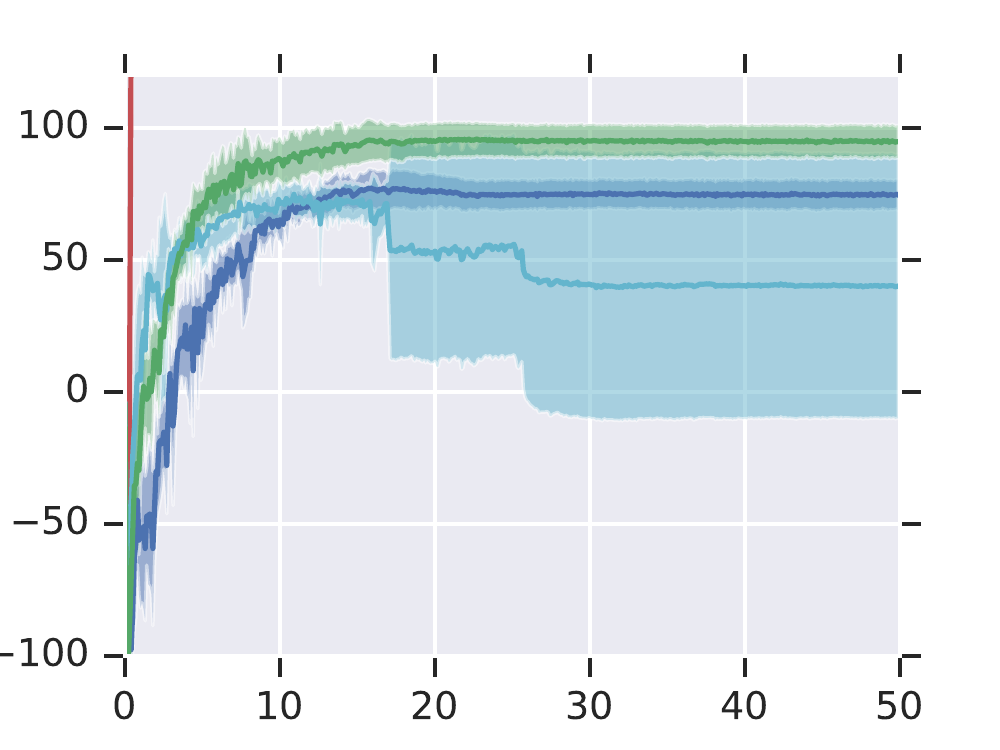}} &
      \subfloat[\scriptsize HalfCheetah-Safe, Constraint]{\label{fig:hc-con}\includegraphics[trim=0mm 0mm 0mm 0mm,clip,width=0.22\textwidth, height=2.2cm,keepaspectratio=false]{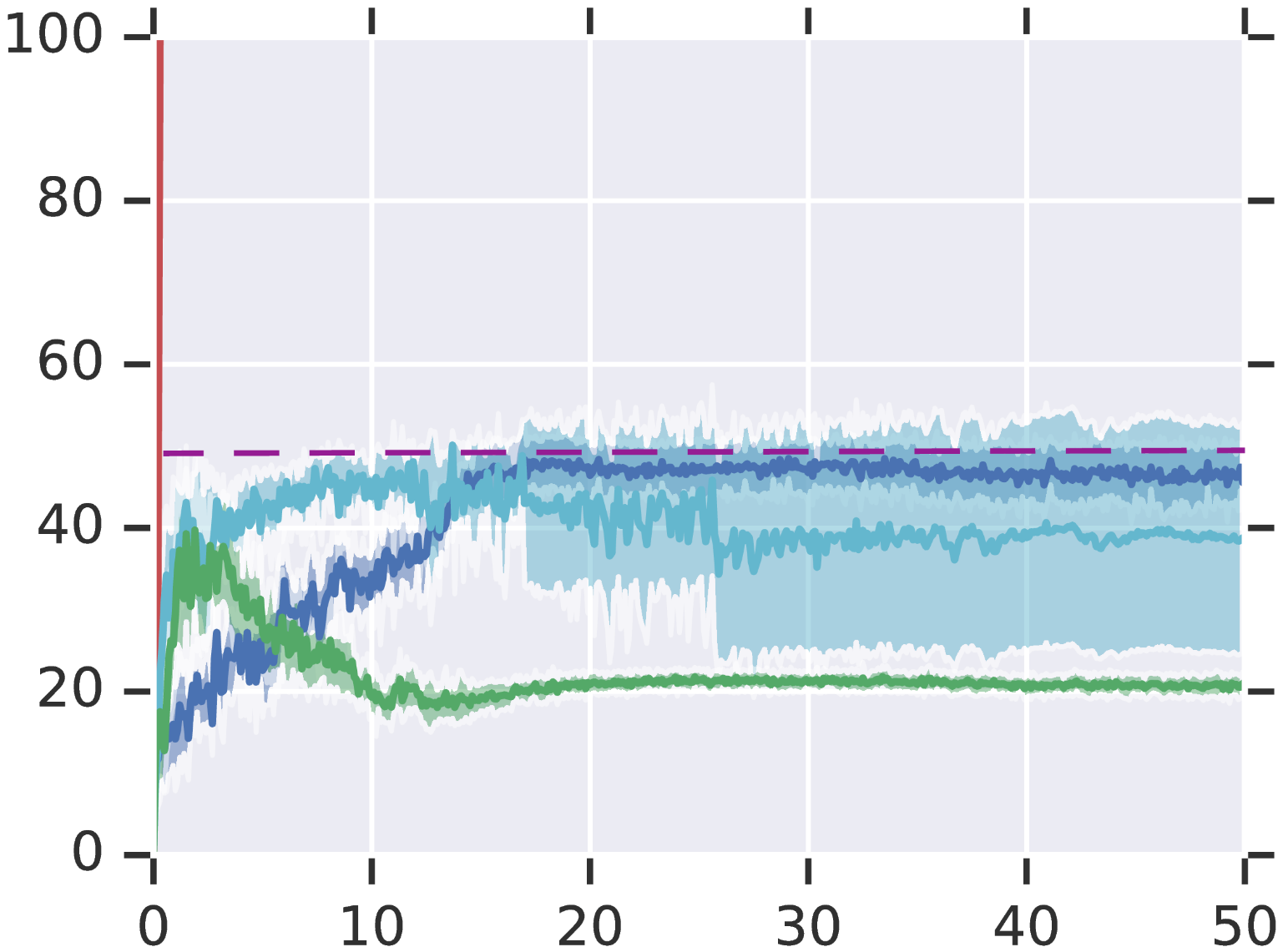}} \\
      \subfloat[\scriptsize Point-Gather, Return]{\label{fig:pg-ret}\includegraphics[trim=3mm 3mm 7mm 3mm,clip,width=0.22\textwidth, height=2.2cm,keepaspectratio=false]{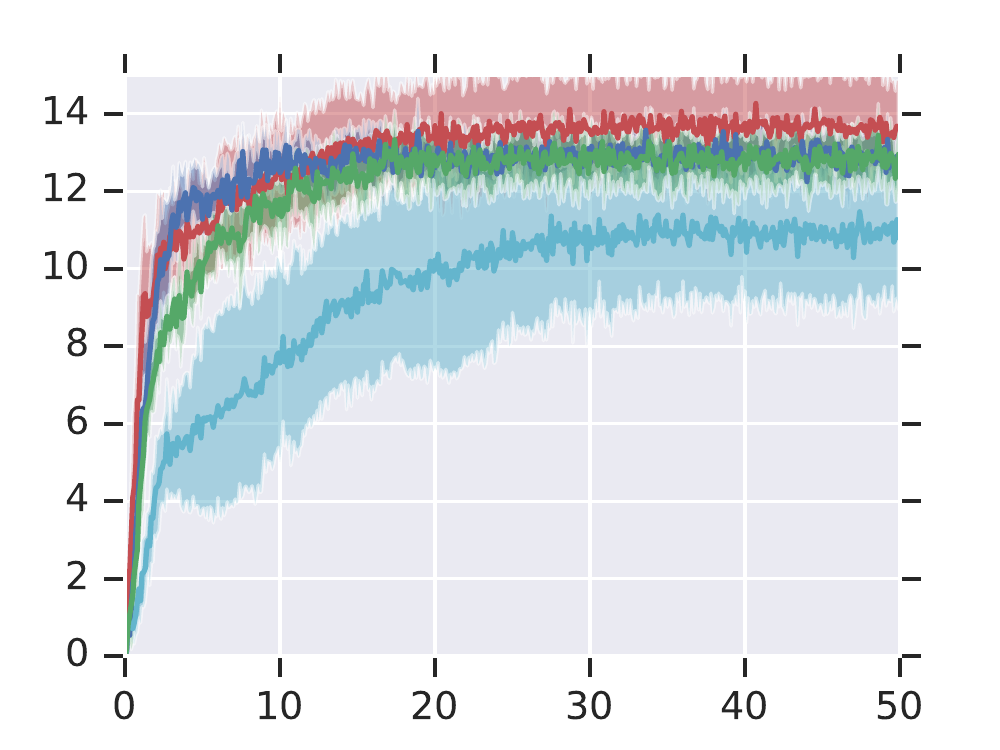}} &
      \subfloat[\scriptsize Point-Gather, Constraint]{\label{fig:pg-con}\includegraphics[trim=0mm 0mm 0mm 0mm,clip,width=0.22\textwidth, height=2.2cm,keepaspectratio=false]{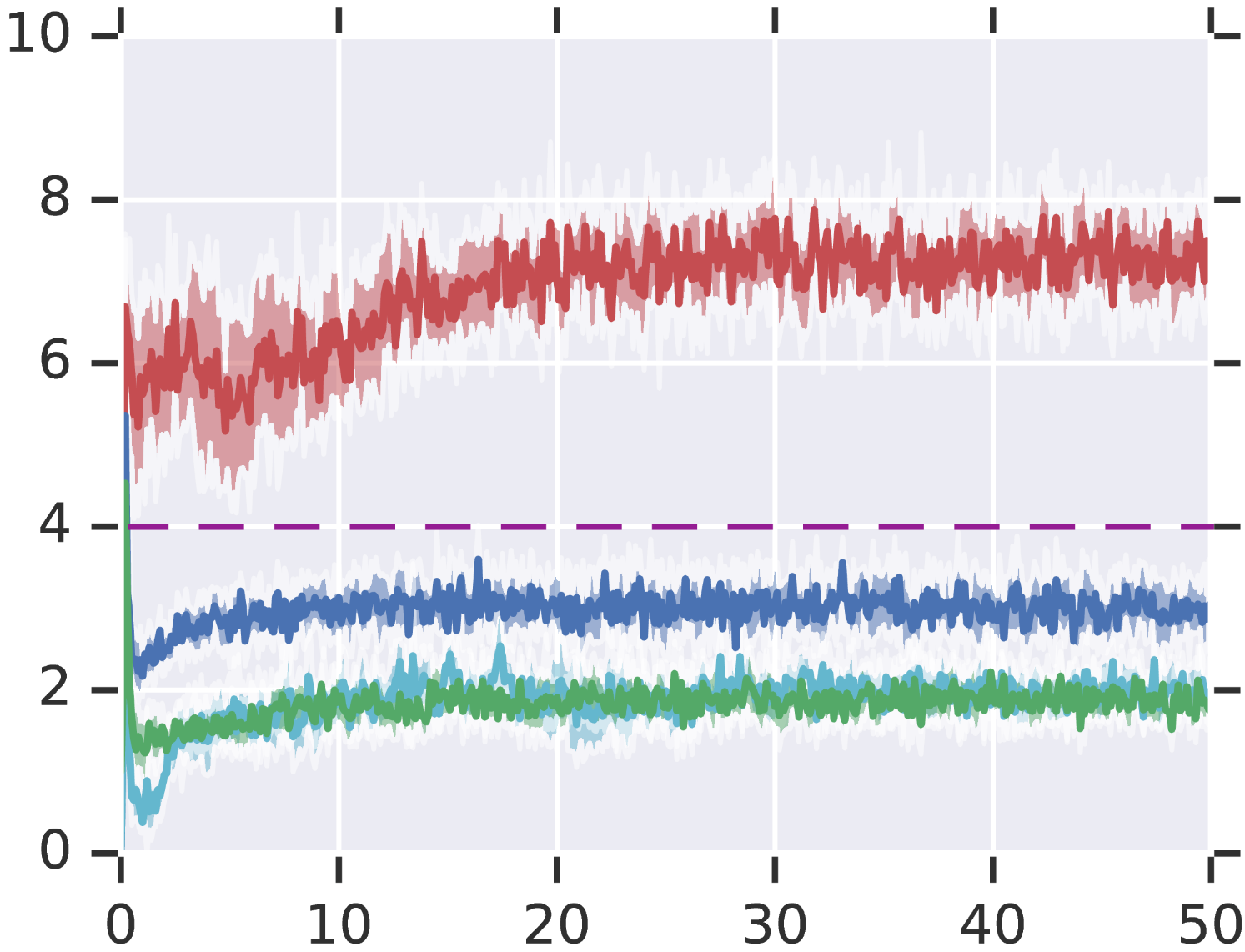}} \\

      \subfloat[\scriptsize Ant-Gather, Return]{\label{fig:ag-ret}\includegraphics[trim=3mm 3mm 7mm 3mm,clip,width=0.22\textwidth, height=2.2cm,keepaspectratio=false]{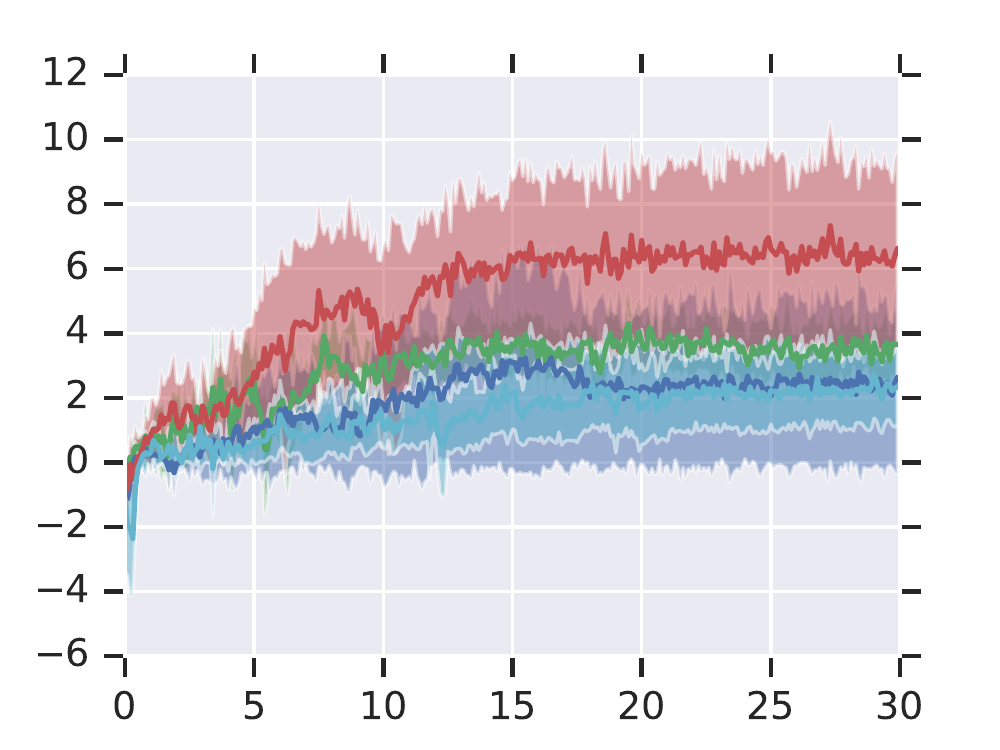}} &
      \subfloat[\scriptsize Ant-Gather, Constraint]{\label{fig:ag-con}\includegraphics[trim=0mm 0mm 0mm 0mm,clip,width=0.22\textwidth, height=2.2cm,keepaspectratio=false]{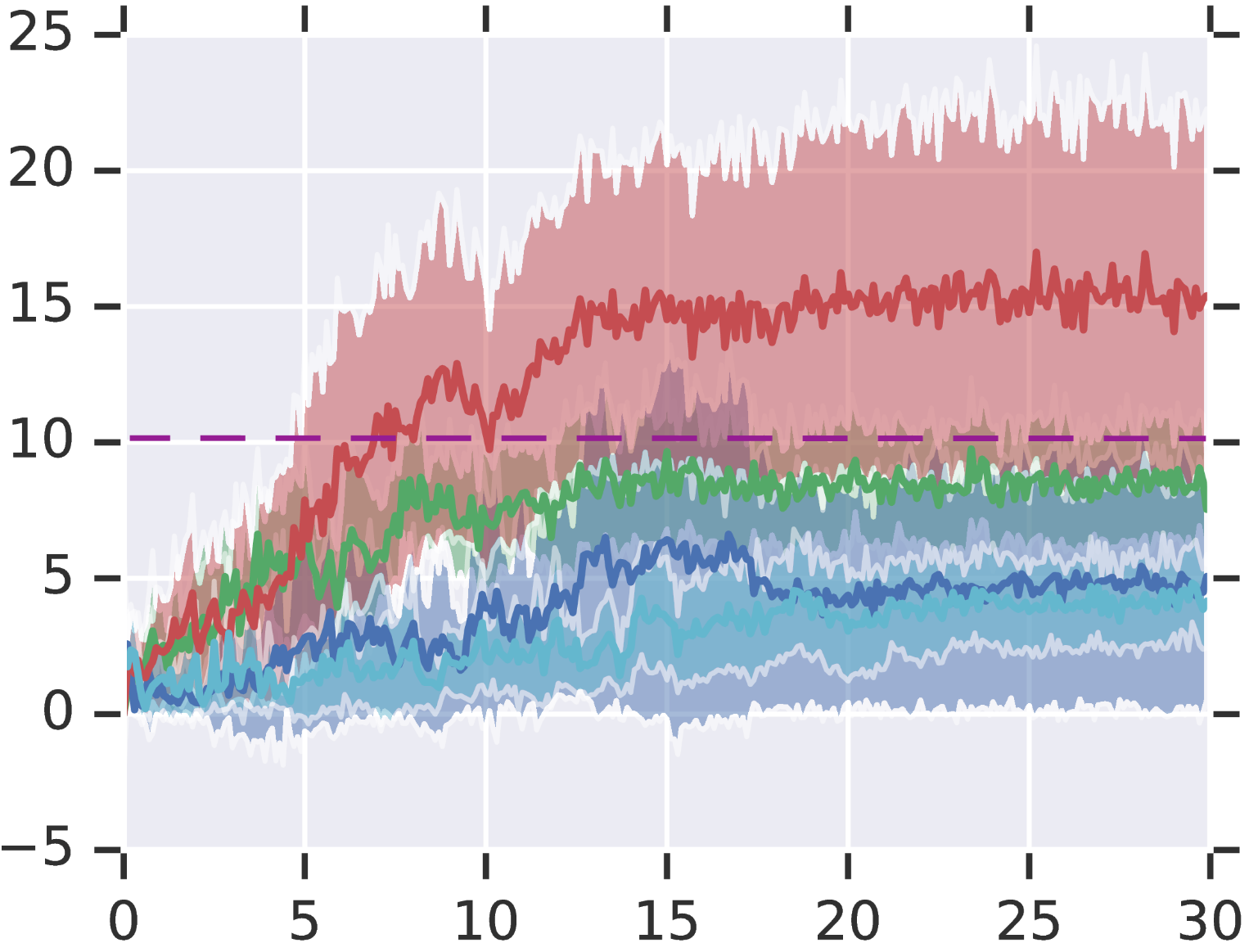}} \\      
      \subfloat[\scriptsize Point-Circle, Return]{\label{fig:pc-ret}\includegraphics[trim=3mm 3mm 7mm 3mm,clip,width=0.22\textwidth, height=2.2cm,keepaspectratio=false]{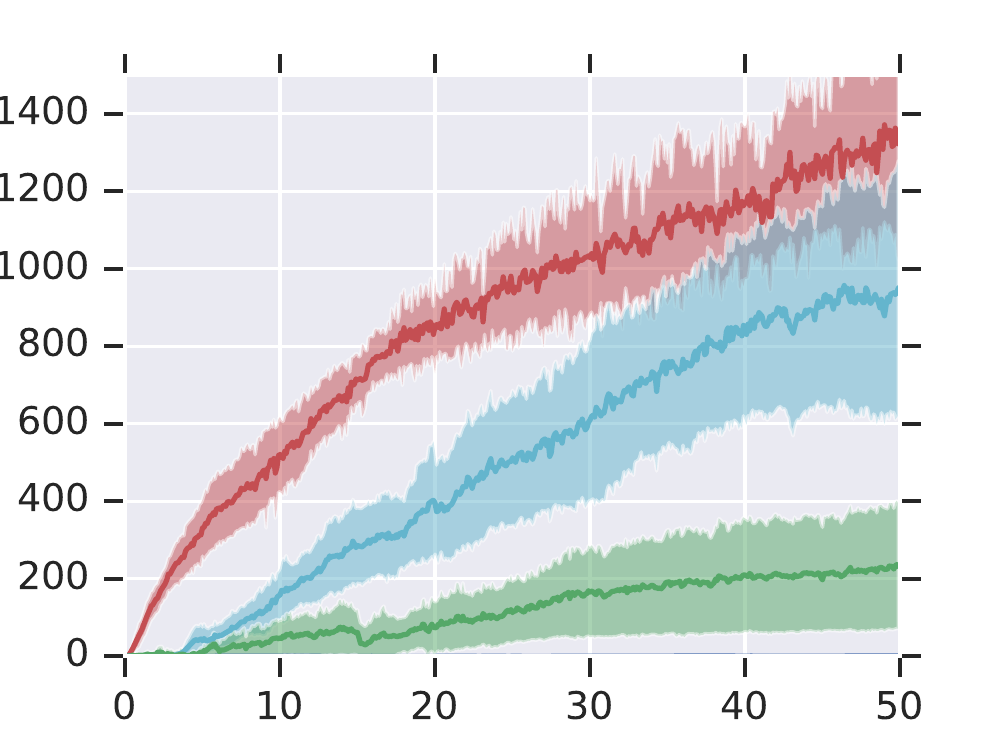}} &
      \subfloat[\scriptsize Point-Circle, Constraint]{\label{fig:pc-con}\includegraphics[trim=0mm 0mm 0mm 0mm,clip,width=0.22\textwidth, height=2.2cm,keepaspectratio=false]{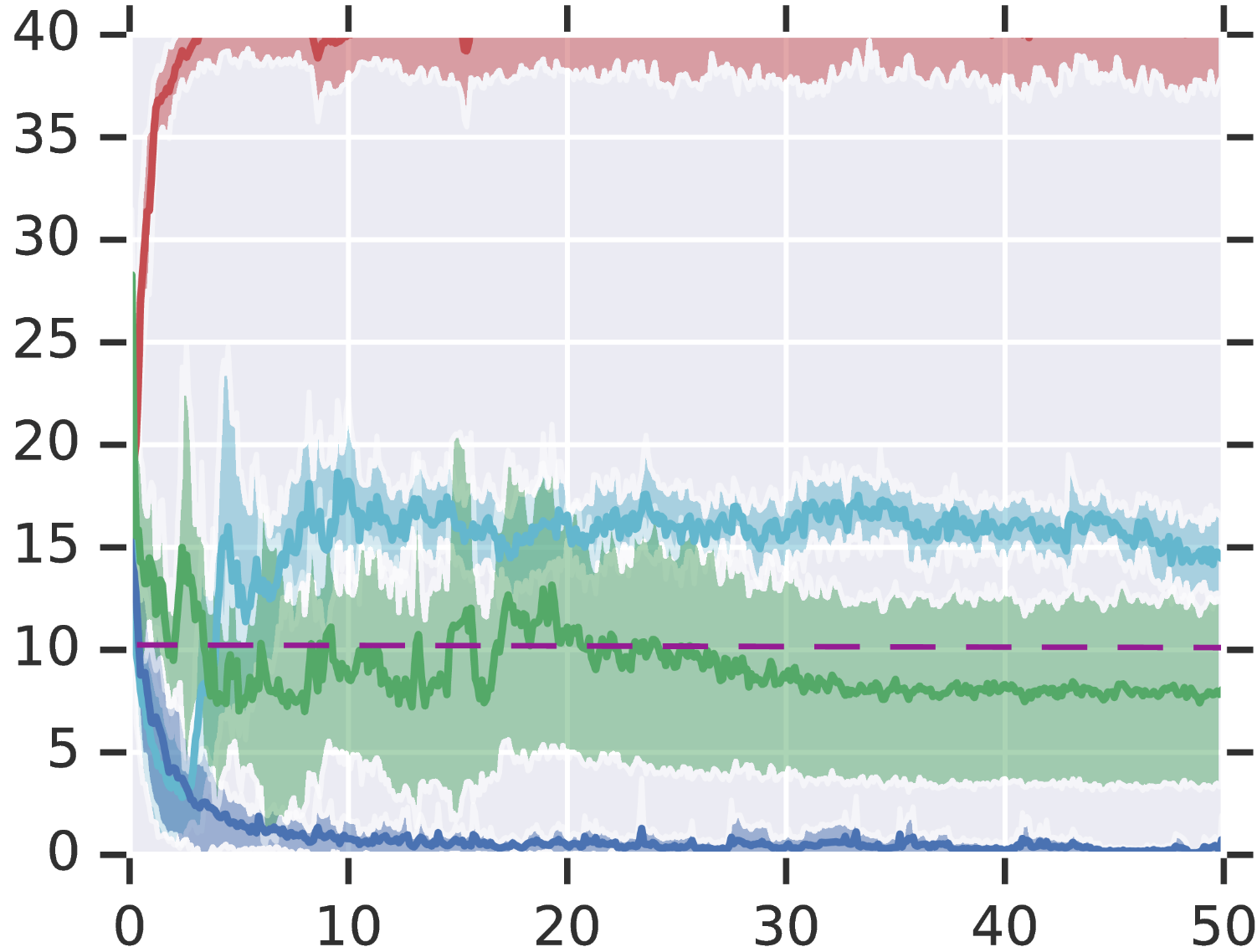}} \\            
  \end{tabular}
\caption{\footnotesize 
DDPG (red), DDPG-Lagrangian (cyan), SDDPG (blue), DDPG $a$-projection (green) on HalfCheetah-Safe  and Point-Gather. Ours (SDDPG, SDDPG $a$-projection) perform stable and safe learning, although the dynamics and cost functions are not known, control actions are continuous, and deep function approximations are necessary. Unit of x-axis is in thousands of episodes. Shaded areas represent the $1$-SD confidence intervals (over $10$ random seeds). The dashed purple line represents the constraint limit.
}
\vspace{-0.15in}
\label{fig_ddpg}
\end{figure}

We empirically evaluate the Lyapunov-based PG algorithms to assess: (i) the performance in terms of cost and safety during training, and (ii) robustness with respect to constraint violations in the presence of function approximation errors. 
To that end, we design three interpretable experiments in simulated robot locomotion continuous control tasks using the MuJoCo simulator \citep{mujoco}. The tasks notions of safety are motivated by physical constraints: (i)\underline{HalfCheetah-Safe}: The HalfCheetah agent is rewarded for running, but its speed is limited for stability and safety;
(ii) \underline{Point-Circle}: The Point agent is rewarded for running in a wide
circle, but is constrained to stay within a safe region defined by $|x|\le x_{\text{lim}}$~\citep{achiam2017constrained};
(iii) \underline{Point-Gather \& Ant-Gather}: Point or Ant Gatherer agent, is rewarded for collecting target objects in a terrain map, while being constrained to avoid bombs~\citep{achiam2017constrained}. Visualizations of these tasks as well as more details of the network architecture used in training the algorithms are given in Appendix \ref{appendix:exp}.

We compare the presented methods with two state-of-the-art unconstrained reinforcement learning algorithms, DDPG \citep{lillicrap2015continuous} and PPO \citep{schulman2017proximal}, and two constrained methods, Lagrangian approach with optimized hyper-parameters for fairness (Appendix \ref{appendix:lag}) and on-policy CPO algorithm \cite{achiam2017constrained}.
The original CPO is based on TRPO \citep{schulman2015trust}. We use its PPO alternative (which coincides with the SPPO algorithm derived in Section 4.1) as the safe RL baseline. SPPO preserves the essence of CPO by adding the first order constraint and the relative entropy regularization to the policy optimization problem. The main difference between CPO and SPPO is that the latter \emph{does not} perform backtracking line-search in learning rate. The decision to compare with SPPO instead of CPO is 1) to avoid the additional computational complexity of line-search in TRPO, while maintaining the performance of PG using the popular PPO algorithm, 2) to have a back-propagatable version of CPO, and 3) to have a fair comparison with other back-propagatable safe RL algorithms, such as the DDPG and safety layer counterparts.

\begin{figure}[tb]
\begin{small}
\centering
  \begin{tabular}{cc}
      \subfloat[\scriptsize HalfCheetah-Safe, Return]{\label{fig:ppo-hc-ret}\includegraphics[trim=3mm 3mm 7mm 3mm,clip,width=0.22\textwidth, height=2.2cm,keepaspectratio=false]{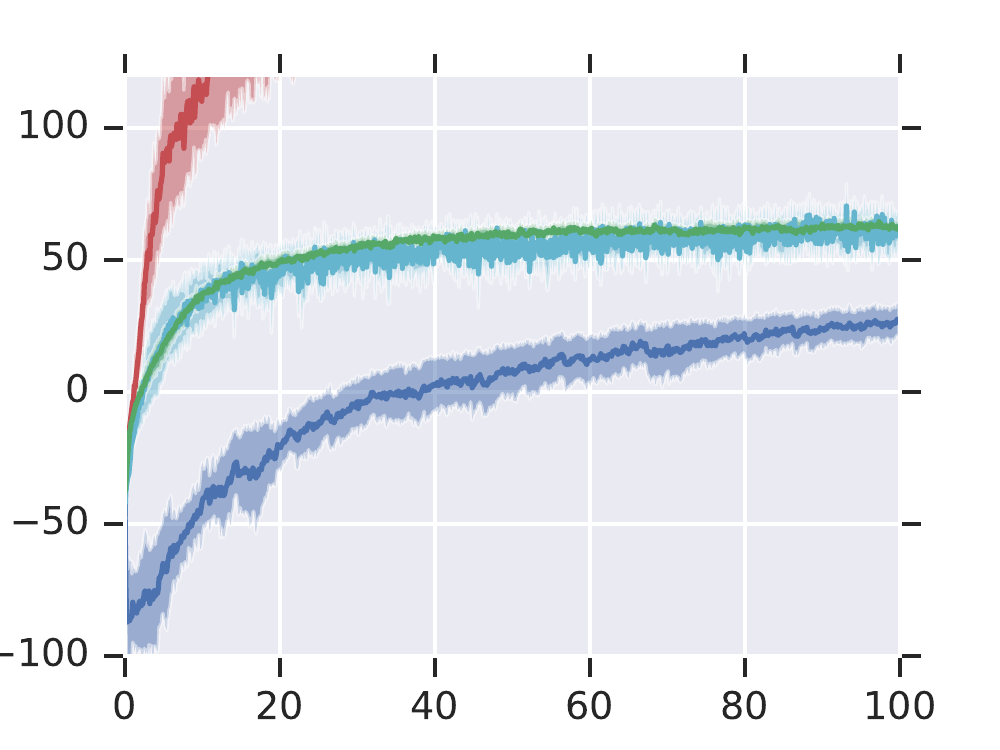}} &
      \subfloat[\scriptsize HalfCheetah-Safe, Constraint]{\label{fig:ppo-hc-con}\includegraphics[trim=3mm 3mm 7mm 3mm,clip,width=0.22\textwidth, height=2.2cm,keepaspectratio=false]{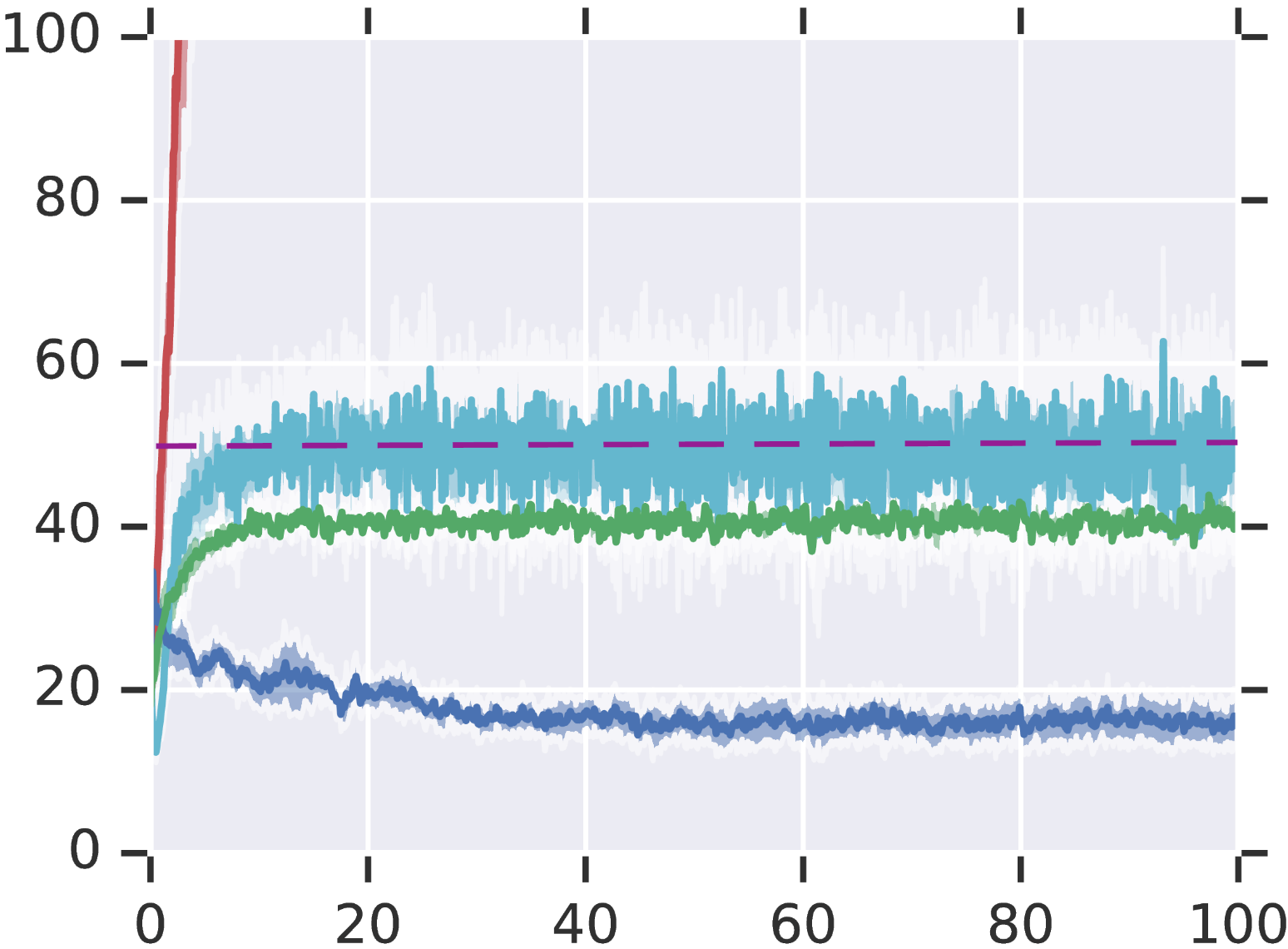}} \\
      \subfloat[\scriptsize Point-Gather, Return]{\label{fig:ppo-pg-ret}\includegraphics[trim=3mm 3mm 7mm 3mm,clip,width=0.22\textwidth, height=2.2cm,keepaspectratio=false]{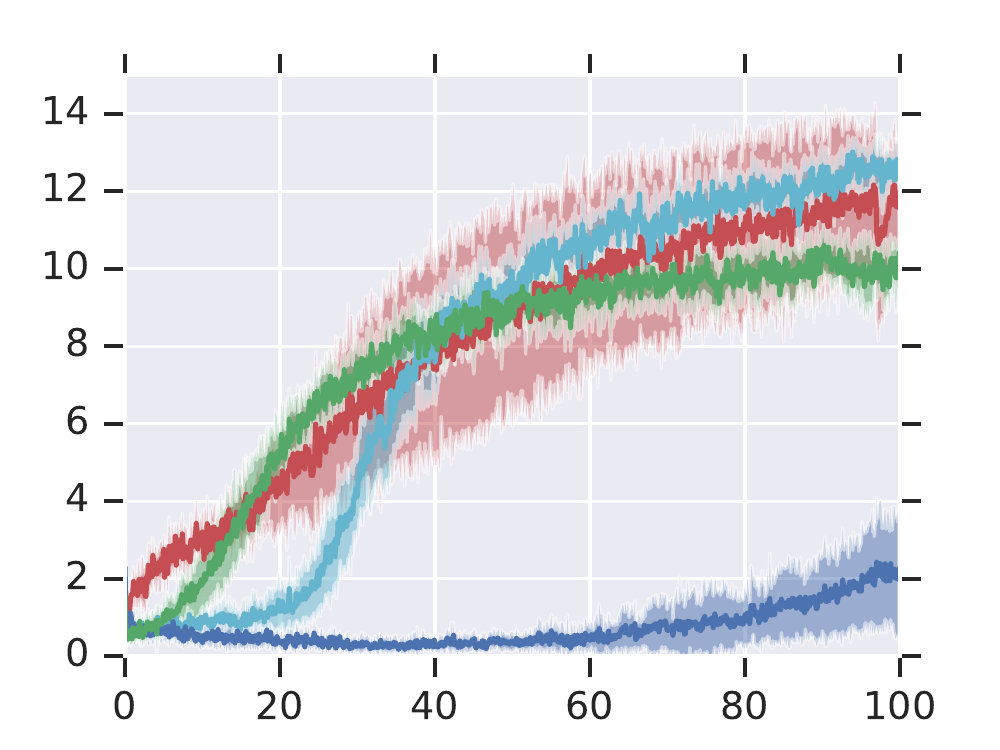}} &
      \subfloat[\scriptsize Point-Gather, Constraint]{\label{fig:ppo-pg-con}\includegraphics[trim=3mm 3mm 7mm 3mm,clip,width=0.22\textwidth, height=2.2cm,keepaspectratio=false]{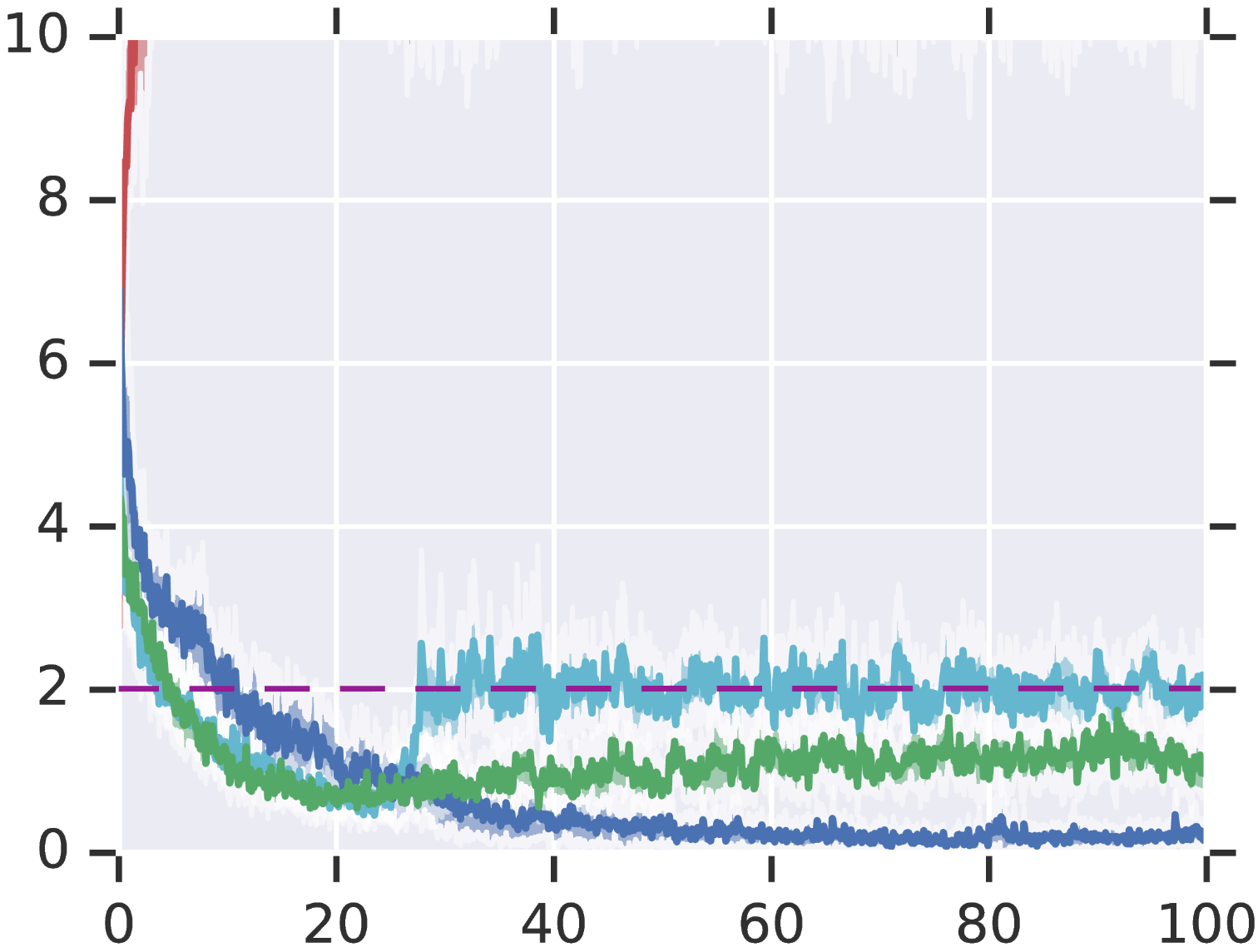}} \\
      
      \subfloat[\scriptsize Ant-Gather, Return]{\label{fig:ppo-ag-ret}\includegraphics[trim=3mm 3mm 7mm 3mm,clip,width=0.22\textwidth, height=2.2cm,keepaspectratio=false]{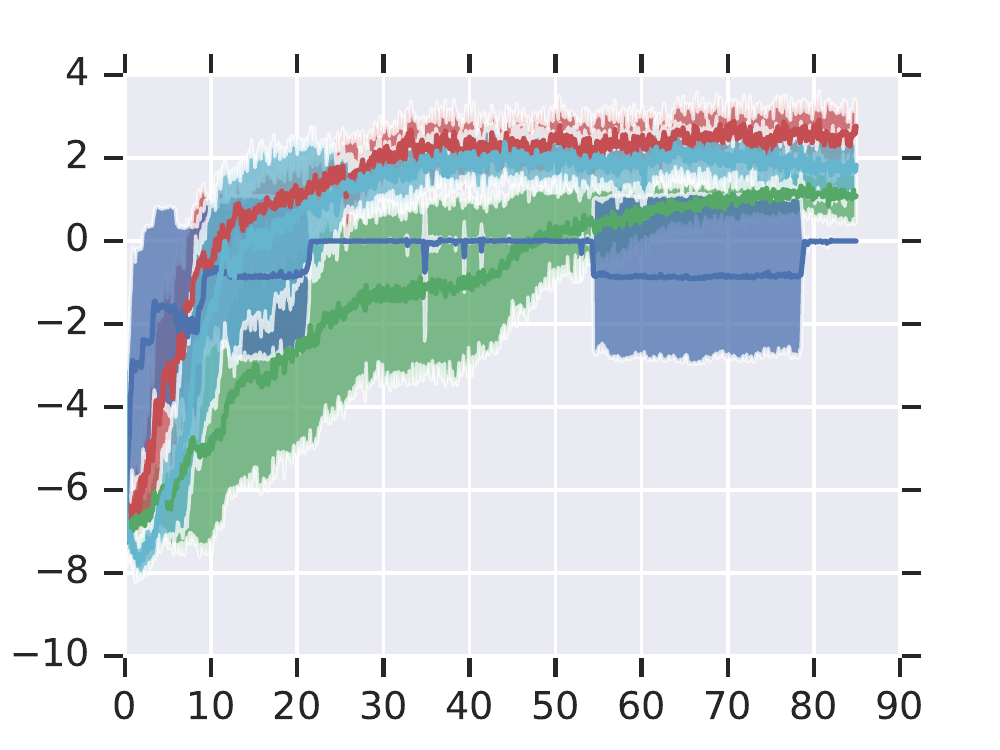}} &
      \subfloat[\scriptsize Ant-Gather, Constraint]{\label{fig:ppo-ag-con}\includegraphics[trim=3mm 3mm 7mm 3mm,clip,width=0.22\textwidth, height=2.2cm,keepaspectratio=false]{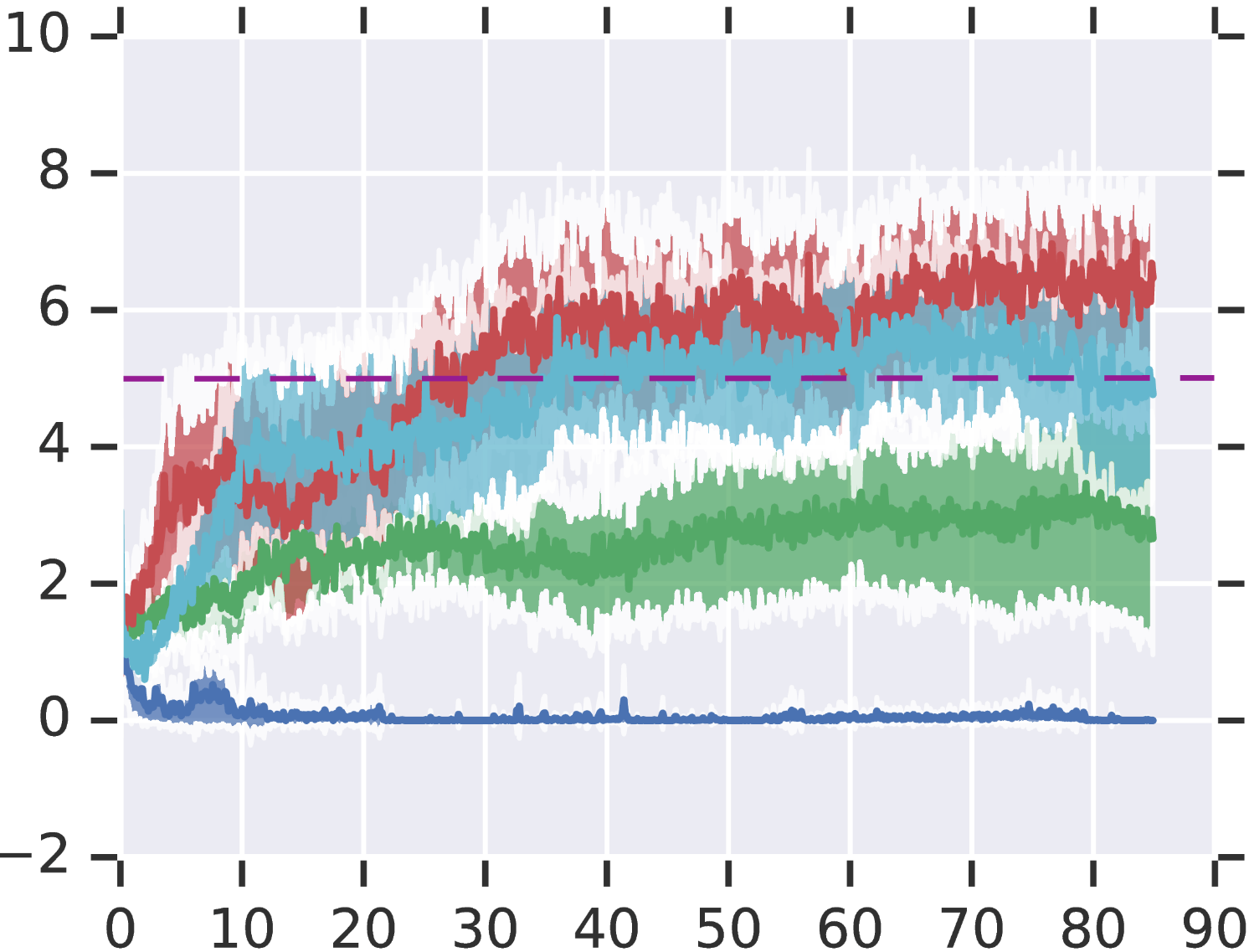}} \\
      
    \subfloat[\scriptsize Point-Circle, Return]{\label{fig:ppo-pc-ret}\includegraphics[trim=3mm 3mm 7mm 3mm,clip,width=0.22\textwidth, height=2.2cm,keepaspectratio=false]{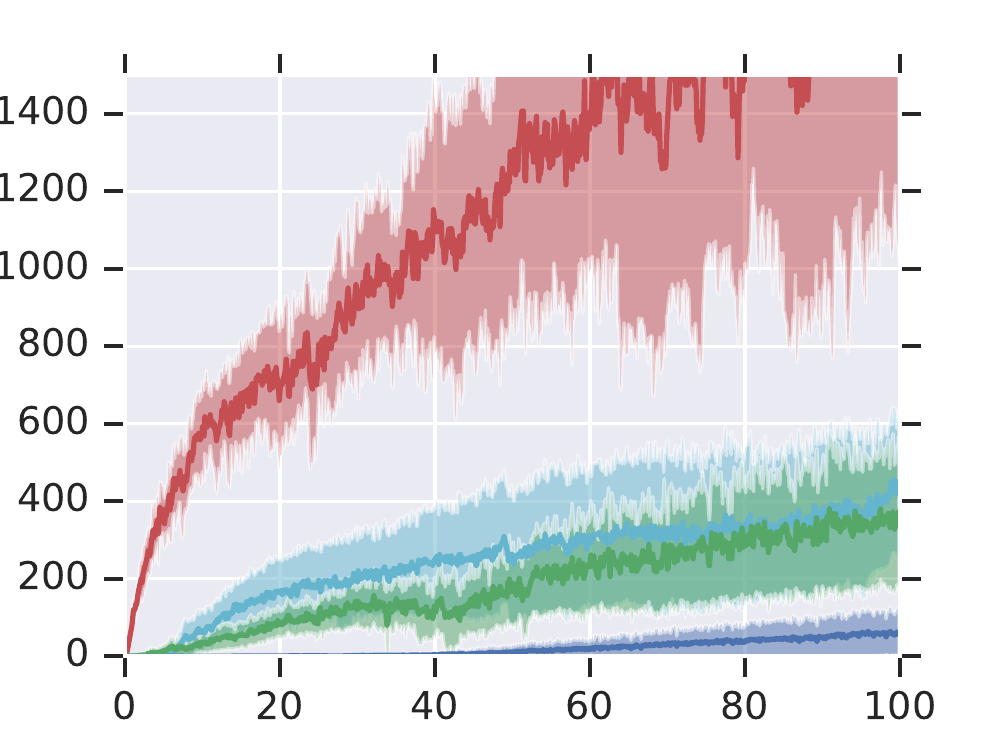}} &
      \subfloat[\scriptsize Point-Circle, Constraint]{\label{fig:ppo-pc-con}\includegraphics[trim=0mm 0mm 0mm 0mm,clip,width=0.22\textwidth, height=2.2cm,keepaspectratio=false]{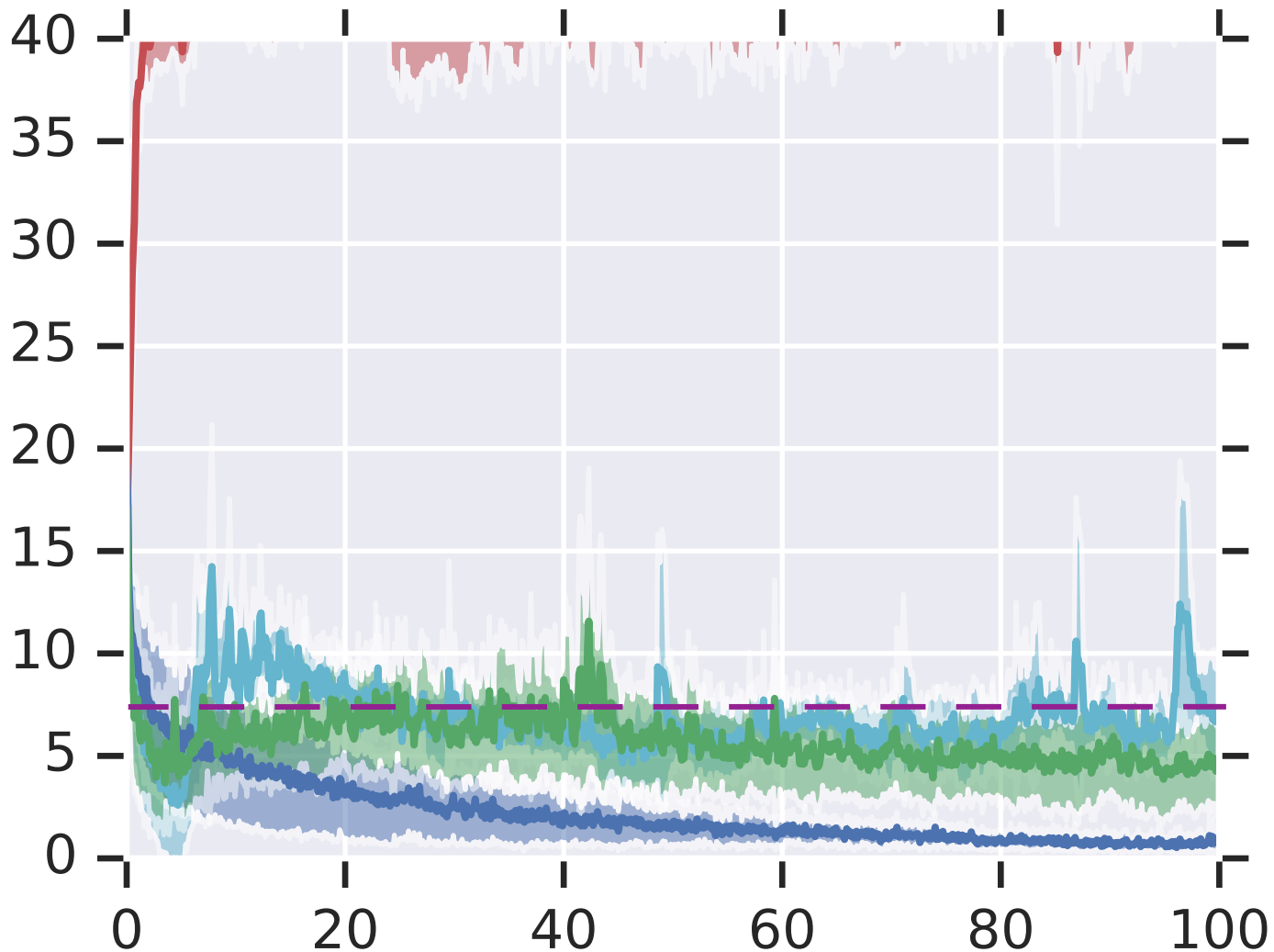}} \\
  \end{tabular}
\caption{\footnotesize
PPO (red), PPO-Lagrangian (cyan), SPPO (blue), SPPO $a$-projection (green) on HalfCheetah-Safe and Point-Gather. Ours (PPO, SPPO $a$-projection) perform stable and safe learning, when the dynamics and cost functions are not known, control actions are continuous, and deep function approximations are necessary. 
}
\label{fig_ppo}
\vspace{-0.15in}
\end{small}
\end{figure}

\textit{Comparisons with baselines:} The Lyapunov-based PG algorithms are stable in learning and all methods converge to feasible
policies with reasonable performance (Figures \ref{fig:hc-ret}, \ref{fig:pg-ret}, \ref{fig:ag-ret}, \ref{fig:pc-ret}, \ref{fig:ppo-hc-ret}, \ref{fig:ppo-pg-ret}, \ref{fig:ppo-ag-ret}, \ref{fig:ppo-pc-ret}). In contrast, when examining the constraint violation (Figures \ref{fig:hc-con}, \ref{fig:pg-con}, \ref{fig:ag-con}, \ref{fig:pc-con}, \ref{fig:ppo-hc-con}, \ref{fig:ppo-pg-con}, \ref{fig:ppo-ag-con}, \ref{fig:ppo-pc-ret}), the Lyapunov-based PG algorithms quickly stabilize the constraint cost to be below the threshold, while the unconstrained DDPG and PPO agents violate the constraints in these environments, and the the Lagrangian approach tends to jiggle around the constrain threshold. Furthermore it is worth-noting that the Lagrangian approach can be sensitive to the
initialization of the Lagrange multiplier $\lambda_0$. If $\lambda_0$ is too large, it would make policy updates overly conservative, while if $\lambda_0$ is too small then constraint violation will be more pronounced. Without further knowledge about the environment, here we treat $\lambda_0$ as a hyper-parameter and optimize it via grid-search. See Appendix \ref{appendix:exp} for more detail. 

\textit{$a$-projection vs. $\theta$-projection:} In many cases the $a$-projection (DDPG and PPO $a$-projections) converges faster and has lower constraint violation than its $\theta$-projection counterpart (SDDPG, SPPO). 
This corroborates with the hypothesis that the $a$-projection approach is less conservative during policy updates than the $\theta$-projection approach (which is what CPO is based on) and generates smoother gradient updates during end-to-end training, resulting in more effective learning than CPO ($\theta$-projection).

\textit{DDPG vs. PPO:} Finally, in most experiments (HalfCheetah, PointGather, and AntGather) the DDPG algorithms tend to have faster learning than the PPO counterpart, while the PPO algorithms have better control on constraint violations (which are able to satisfy lower constraint thresholds). The faster learning behavior is potentially due to the improved data-efficiency when using off-policy samples in PG updates, however the covariate-shift in off-policy data makes tight constraint control more challenging. 

\vspace{-0.05in}
\section{Safe Policy Gradient for Robot Navigation}\label{sec:robot}
\vspace{-0.05in}
\begin{figure}[tb]
  \begin{tabular}{c}
    \subfloat[\scriptsize Noisy Lidar observation in a corridor]{\label{fig:lidar}\includegraphics[trim=0mm 15mm 0mm 15mm,clip,width=0.42\textwidth, height=0.7cm,keepaspectratio=false]{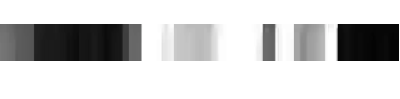}} \\
    \vspace{-0.1in}
    \subfloat[\scriptsize SDDPG for point to point task]
    {\includegraphics[trim=0mm 10mm 0mm 0mm,clip,width=0.42\textwidth, height=3.0cm,keepaspectratio=false]{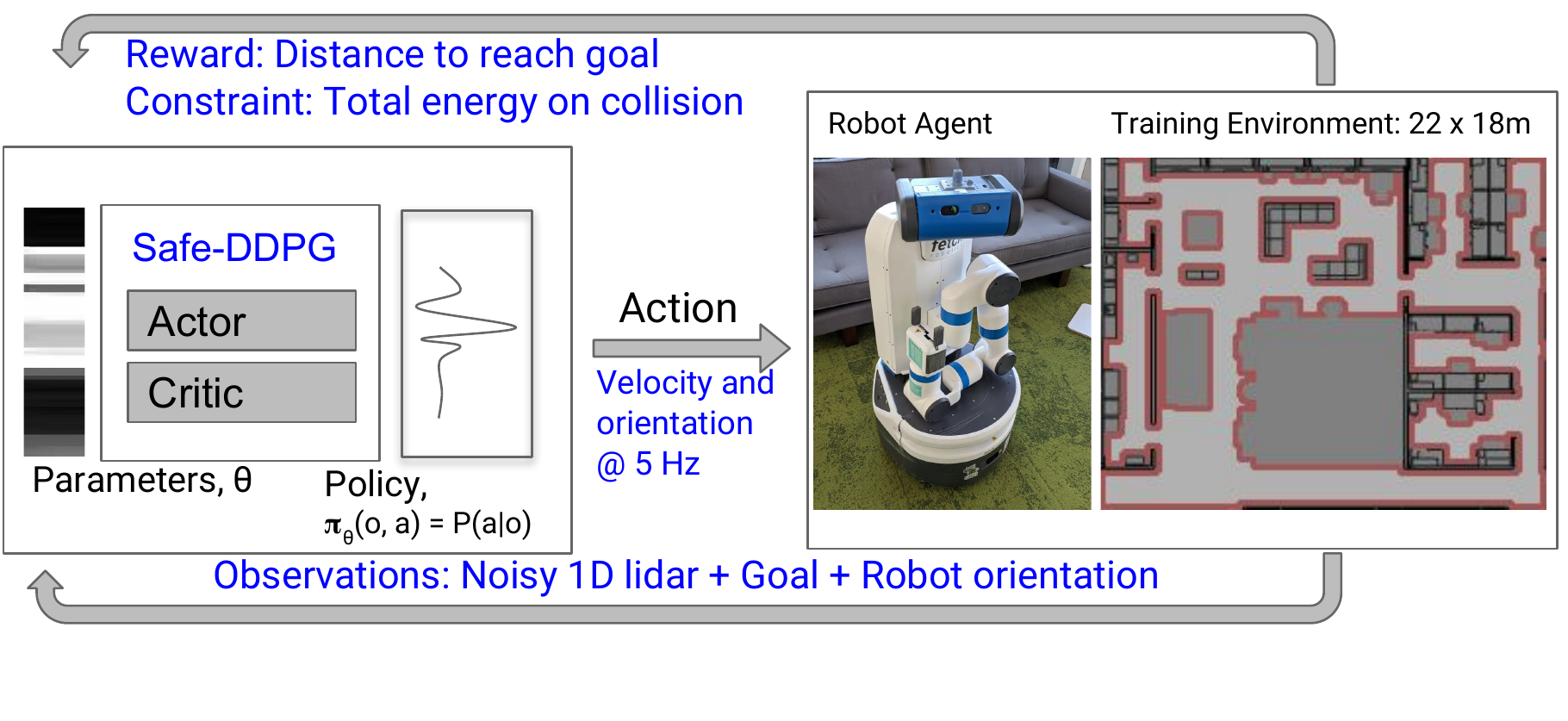}\label{fig:sddpg}}
  \end{tabular}
  \caption{\footnotesize Robot navigation task details. \label{fig:fetchFun}}
\end{figure}

We now evaluate the safe policy optimization on a real robot task -- point to point (P2P) navigation \citep{autorl} -- where a noisy differential drive robot with limited sensors (Fig. \ref{fig:lidar}), is required to navigate to a goal outside of its visual field of view while avoiding collisions with obstacles. 
The agent's observations consist of the relative goal position, the relative goal velocity, and the Lidar measurements (Fig. \ref{fig:lidar}). The actions are the linear and angular velocity vector at the robot's center of the mass. The transition probability captures the noisy differential drive robot dynamics, whose exact formulation is not known to the robot. The robot must navigate to arbitrary goal positions collision-free and without memory of the workspace topology.

Here the CMDP is non-discounting and has a fixed horizon. We reward the agent for reaching the goal, which translates to an immediate cost that measures the relative distance to goal.
To measure the impact energy of obstacle collisions, we impose an immediate constraint cost to account for the speed during collision, with a constraint threshold $d_0$ that characterizes the agent's maximum tolerable collision impact energy to any objects. This type of constraint allows the robot to touch the obstacle (such as walls) but prevent it from ramming into any objects. Under this CMDP framework (Fig. \ref{fig:sddpg}), the main goal is to train a policy $\pi^*$ that drives the robot along the shortest path to the goal and to limit the total impact energy of obstacle collisions. Furthermore, we note that due to limited data, in practice intermediate point-to-point policies are deployed on the real-world robot to collect more samples for further training. Therefore, guaranteeing safety during training is critical in this application.
Descriptions about the robot navigation problem, including training and evaluation environments are in Appendix \ref{appendix:robot}.

\begin{figure}[tb]
  \begin{tabular}{cc}
    \subfloat[\scriptsize Navigation, Mission Success $\%$]{\label{fig:robot-ret}\includegraphics[width=0.21\textwidth, height=3.0cm,keepaspectratio=false]{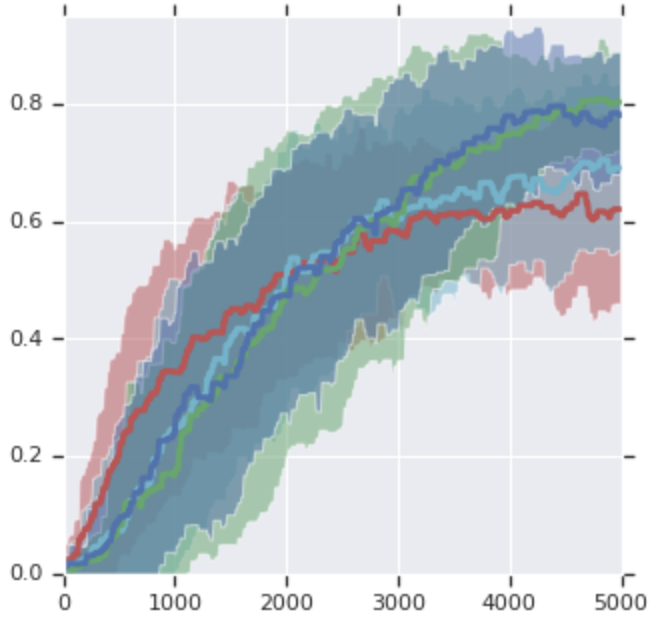}} &
    \subfloat[\scriptsize Navigation, Constraint]{\includegraphics[width=0.21\textwidth, height=3.0cm,keepaspectratio=false]{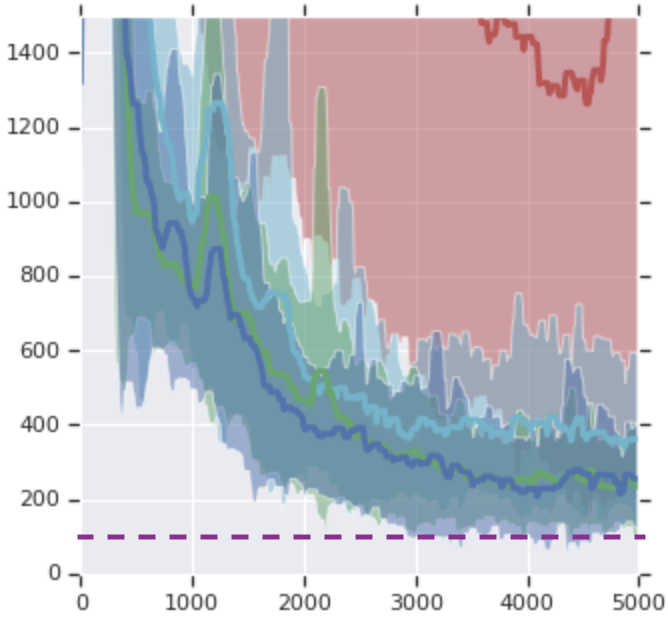}\label{fig:robot-con}}
    \vspace{-0.1in}
  \end{tabular}
   \caption{\footnotesize 
DDPG (red), DDPG-Lagrangian (cyan), SDDPG (blue), DDPG $a$-projection (green) on Robot Navigation. Ours (SDDPG, SDDPG $a$-projection) balance between reward and constraint learning. Unit of x-axis is in thousands of steps. The shaded areas represent the $1$-SD confidence intervals (over $50$ runs). The dashed purple line represents the constraint limit.
\label{fig:robot_ddpg}}
\end{figure}

\begin{figure}[tb]
  \begin{tabular}{ccc}
    \subfloat[\scriptsize Lagrangian policy]{\label{fig:lag-nav}\includegraphics[width=0.135\textwidth]{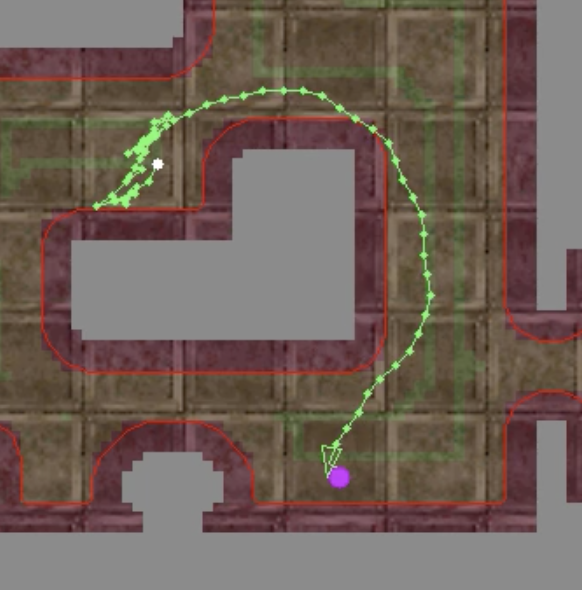}}\quad &
    \subfloat[\scriptsize SDDPG ($a$-proj.)]{\includegraphics[width=0.135\textwidth]{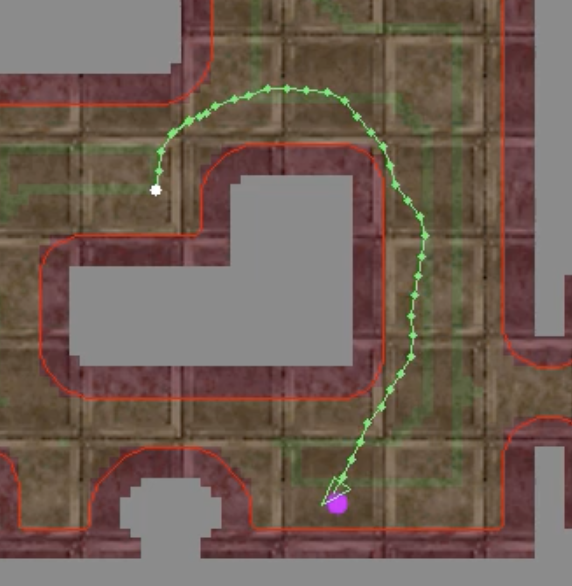}\label{fig:lyap-nav}}\quad &
    \subfloat[\scriptsize SDDPG ($a$-proj.) on robot]{\includegraphics[width=0.13\textwidth]{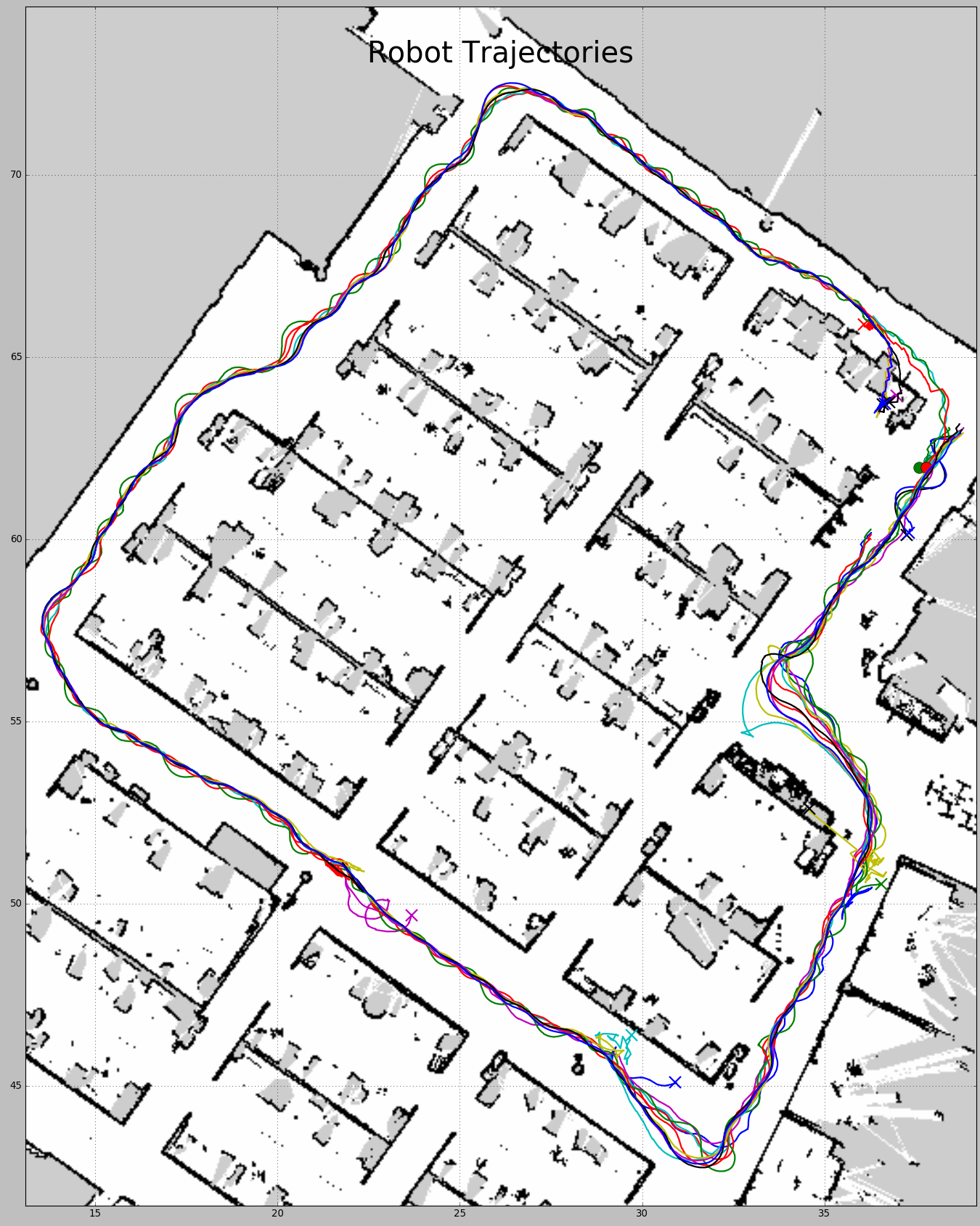}\label{fig:on-robot}}    
  \end{tabular}
      \vspace{-0.1in}
   \caption{\footnotesize 
Navigation routes of two policies on a similar setup (a) and (b). Log of on-robot experiments (c). Larger version in Appendix \ref{appendix:robot} and the video is available in the supplementary materials.
\label{fig:robot_ddpg_2}}
\end{figure}

\textit{Experimental Results:} 
We evaluate the learning algorithms in terms of average mission success percentage and constraint control. The task is successful if the robot reaches the goal before the constraint threshold (total energy of collision) is exhausted, and the success rate is averaged over $100$ evaluation episodes with random initialization. While all methods converge to policies with reasonable performance, Figure \ref{fig:robot-ret} and \ref{fig:robot-con} shows that the Lyapunov-based PG algorithms have higher success rates, due to their robust abilities of controlling the total constraint, as well minimizing the distance to goal. Although the unconstrained method often yields a lower distance to goal, it violates the constraint more frequently and thus leads to a lower success rate. Furthermore, note that the Lagrangian approach is less robust to initialization of parameters, and therefore it generally has lower success rate and higher variability than the Lyapunov-based methods. Unfortunately due to function approximation error and stochasticity of the problem, all the algorithms converged pre-maturely with constraints above the threshold. One reason is due to the constraint threshold ($d_0=100$) being overly-conservative. In real-world problems guaranteeing constraint satisfaction is more challenging than maximizing return, and that usually requires much more training data. 
Finally, Figures \ref{fig:lag-nav} and \ref{fig:lyap-nav} illustrate the navigation routes of two policies. On similar goal configurations, the Lagrangian method tends to zigzag and has more collisions, while the Lyapunov-based algorithm (SDDPG)  chooses a safer path to reach the goal.

\begin{figure}[h!]
  \begin{tabular}{cc}
    \subfloat[\scriptsize Navigation, Mission Success $\%$]{\label{fig:goal-successt}\includegraphics[width=0.225\textwidth,trim=3mm 3mm 3mm 3mm,clip, height=2.5cm,keepaspectratio=false]{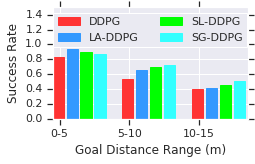}} &
    \subfloat[\scriptsize Navigation, Constraint]{\includegraphics[width=0.225\textwidth,trim=3mm 3mm 3mm 3mm,clip, height=2.5cm,keepaspectratio=false]{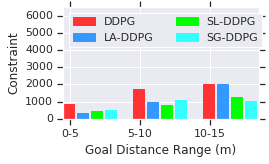}\label{fig:goal-c}}\\
  \end{tabular}
      \vspace{-0.1in}
   \caption{\footnotesize Robot navigation generalization over success rate (a) and constraint satisfaction (b) on a different environment.
\label{fig:robot_gen}}
\end{figure}

Next, we evaluate how well the methods generalize to (i) longer trajectories, and (ii) new environments. P2P tasks are trained in a $22$ by $18$ meters environment (Fig. \ref{fig:robot_task}) with goals placed within $5$ to $10$ meters from the robot initial state,
Figure \ref{fig:robot_gen} depicts the results evaluations, averaged over $100$ trials, on P2P tasks in a much larger evaluation environment ($60$ by $47$ meters) with goals placed up to $15$ meters away from the goal. The success rate of all methods degrades as the goals are further away (Fig. \ref{fig:goal-successt}), and the safety methods ($a$-projection -- SL-DDPG, and $\theta$-projection -- SG-DDPG) outperform unconstrained and Lagrangian (DDPG and LA-DDPG) as the task becomes more difficult. At the same time, our methods retain the lower constraints even when the task is difficult (Fig. \ref{fig:goal-c}).

Finally, we deployed the SL-DDPG policy onto the real Fetch robot \cite{fetch} in an everyday office environment. Figure \ref{fig:on-robot} shows the top down view of the robot log. Robot travelled a total of $500$ meters to complete five repetitions of $12$ tasks, each averaging about $10$ meters to the goal. The experiments included narrow corridors and people walking through the office. The robot robustly avoids both static and dynamic (humans) obstacles coming into its path. We observed additional "wobbling" effects, that was not present in simulation. This is likely due to the wheel slippage at the floor that the policy was not trained for. In several occasions when the robot could not find a clear path, the policy instructed the robot to stay put instead of narrowly passing by the obstacle. This is precisely the safety behavior we want to achieve with the Lyapunov-based algorithms.

\vspace{-0.05in}
\section{Conclusions}
\vspace{-0.05in}
We formulated safe RL as a continuous action CMDP and developed two classes, $\theta$-projection and $a$-projection, of policy optimization algorithms based on Lyapunov functions to learn safe policies with high expected cumulative return. We do so by combining both on and off-policy optimization (DDPG or PPO) with a critic that evaluates the policy and computes its corresponding Lyapunov function. 
We evaluated our algorithms on four high-dimensional simulated robot locomotion tasks and compared them with several baselines. To demonstrate the effectiveness of the Lyapunov-based algorithms in solving real-world problems, we also apply these algorithms to indoor robot navigation, to ensure that the agent's path is optimal and collision-free. Our results indicate that our Lyapunov-based algorithms {\bf 1)} achieve safe learning, {\bf 2)} have better data-efficiency, {\bf 3)} can be more naturally integrated within the standard end-to-end differentiable policy gradient training pipeline, and {\bf 4)} are scalable to tackle real-world problems. Our work is a step forward in deploying RL to real-world problems in which safety guarantees are of paramount importance. 
Future work includes {\bf 1)} further exploration of Lyapunov function properties to improve training stability and safety, {\bf 2)} more efficient use of Lyapunov constraints in constrained policy optimization, and {\bf 3)} extensions of the Lyapunov-approach to the model-based setting to better utilize the agent's dynamics.

\newpage
\bibliography{icml2019}
\bibliographystyle{icml2019}

\newpage
\appendix

\onecolumn

\newpage
\vspace{-0.05in}
\section{The Lyapunov Approach to Solve CMDPs}\label{sec:lyapunov_cmdp}
\vspace{-0.05in}



In this section, we revisit the {\em Lyapunov approach} to solving CMDPs that was proposed by~\citet{chow2018lyapunov} and report the mathematical results that are important in developing our safe policy optimization algorithms. To start, without loss of generality, we assume that we have access to a {\em baseline} feasible policy of~\Eqref{eq:CMDP}, $\pi_B$; i.e. $\pi_B$ satisfies $\mathcal D_{\pi_B}(x_0)\le d_0$. We define a set of Lyapunov functions w.r.t.~initial state $x_0\in\mathcal X$ and constraint threshold $d_0$ as 
\[
\mathcal L_{\pi_B}(x_0,d_0)\!=\!\{L:\mathcal X\rightarrow\!\mathbb R_{\geq 0}: T_{\pi_{B},d}[L](x)\!\leq\! L(x),\forall x\in\mathcal X;\, L(x_0)\leq d_0 \},
\]
and call the constraints in this feasibility set the {\em Lyapunov constraints}. For any arbitrary Lyapunov function $L\in\mathcal L_{\pi_{B}}(x_0,d_0)$, we denote by 
\[
\mathcal F_{L}(x)=\left\{\pi(\cdot|x)\in\Delta:T_{\pi,d}[L](x)\leq\! L(x)\right\}
\]
the set of $L$-induced Markov stationary policies. Since $T_{\pi,d}$ is a contraction mapping~\citep{bertsekas2005dynamic}, any $L$-induced policy $\pi$ has the property $\mathcal D_{\pi}(x)=\lim_{k\rightarrow\infty}T^k_{\pi,d}[L](x)\leq L(x)$, $\forall x\in\mathcal X$. Together with the property that $ L(x_0)\leq d_0$, they imply that any $L$-induced policy is a feasible policy of~\Eqref{eq:CMDP}. However, in general, the set $\mathcal F_{L}(x)$ does not necessarily contain an optimal policy of~\Eqref{eq:CMDP}, and thus it is necessary to design a Lyapunov function (w.r.t.~a baseline policy $\pi_B$) that provides this guarantee. In other words, the main goal is to construct a Lyapunov function $L\in \mathcal L_{\pi_B}(x_0,d_0)$ such that
\begin{equation}
L(x)\geq T_{\pi^*,d}[L](x),\quad\quad L(x_0)\leq d_0.\label{eq:opt_lyap}
\end{equation}
%
\citet{chow2018lyapunov} show in their Theorem~1 that {\bf 1)} without loss of optimality, the Lyapunov function can be expressed as 
\[
L_\epsilon(x):=\mathbb E\left[\sum_{t=0}^\infty \gamma^t (d(x_t)+\epsilon(x_t))\mid \pi_{B},x\right],
\]
where $\epsilon(x)\geq 0$ is some auxiliary constraint cost uniformly upper-bounded by 
\[
\epsilon^*(x):=2 D_{\max}D_{TV}(\pi^*||\pi_{B})(x)/(1-\gamma),
\]
and {\bf 2)} if the baseline policy $\pi_B$  satisfies the condition 
\begin{equation*}
\max_{x\in\mathcal X}\epsilon^*(x)\leq D_{\max}\cdot\min\left\{(1-\gamma)\frac{d_0-\mathcal{D}_{\pi_{B}}(x_0)}{D_{\max}}\;,\;\frac{D_{\max}-(1-\gamma)\overline{\mathcal D}}{D_{\max}+(1-\gamma)\overline{\mathcal D}}\right\},
\end{equation*}
where $\overline{\mathcal D}=\max_{x\in\mathcal X}\max_{\pi} \mathcal D_\pi(x)$ is the maximum constraint cost, then the Lyapunov function candidate $L_{\epsilon^*}$ also satisfies the properties of~\Eqref{eq:opt_lyap}, and thus, its induced feasible policy set $\mathcal F_{L_{\epsilon^*}}$ contains an optimal policy. Furthermore, suppose that the distance between the baseline and optimal policies can be estimated effectively. Using the set of $L_{\epsilon^*}$-induced feasible policies and noting that the \emph{safe} Bellman operator $T[V](x)=\min_{\pi\in\mathcal F_{L_{\epsilon^*}}(x)}T_{\pi,c}[V](x)$
is monotonic and contractive, one can show that $T[V](x)=V(x)$, $\forall x\in\mathcal X$ has a unique fixed point $V^*$, such that $V^*(x_0)$ is a solution of~\Eqref{eq:CMDP}, and an optimal policy can be constructed via greedification, i.e.,~$\pi^*(\cdot|x)\!\in\!\arg\min_{\pi\in\mathcal F_{L_{\epsilon^*}}(x)}T_{\pi,c}[V^*](x)$. This shows that under the above assumption,~\Eqref{eq:CMDP} can be solved using standard dynamic programming (DP) algorithms. While this result connects CMDP with Bellman's principle of optimality, verifying whether $\pi_B$ satisfies this assumption is challenging when a good estimate of $D_{TV}(\pi^*||\pi_{B})$ is not available. To address this issue,~\citet{chow2018lyapunov} propose to approximate $ \epsilon^*$ with an auxiliary constraint cost $\widetilde\epsilon$, which is the \emph{largest} auxiliary cost satisfying the Lyapunov condition $L_{\widetilde \epsilon}(x)\geq T_{\pi_B,d}[L_{\widetilde \epsilon}](x),\,\,\forall x\in\mathcal X$ and the safety condition $L_{\widetilde \epsilon}(x_0)\leq d_0$. The intuition here is that the larger $\widetilde\epsilon$, the larger the set of policies $\mathcal F_{L_{\widetilde \epsilon}}$. Thus, by choosing the largest such auxiliary cost, we hope to have a better chance of including the optimal policy $\pi^*$ in the set of feasible policies. Specifically, $\widetilde \epsilon$ is computed by solving the following linear programming (LP) problem:
%
\begin{equation}
\label{eq:opt_eps_baseline}
\widetilde \epsilon\in\arg\max_{\epsilon:\mathcal X\rightarrow\mathbb R_{\geq 0}}\bigg\{\sum_{x\in\mathcal X}\epsilon(x)\;:\;d_0 - \mathcal D_{\pi_B}(x_0)\geq\mathbf 1(x_0)^\top\Big(I- \gamma\big\{P\big(x'|x,\pi_B(x)\big)\big\}_{x,x'\in\mathcal X}\Big)^{-1}\epsilon\bigg\},
\end{equation}
%
where $\mathbf 1(x_0)$ represents a one-hot vector in which the non-zero element is located at $x=x_0$. When $\pi_B$ is a feasible policy, this problem has a non-empty solution. Furthermore, according to the derivations in~\citet{chow2018lyapunov}, the maximizer of~\eqref{eq:opt_eps_baseline} has the following form: 
\begin{equation*}
\widetilde \epsilon(x)=\frac{\big(d_0 - \mathcal D_{\pi_B}(x_0)\big)\cdot\mathbf 1\{x=\underline x\}}{\mathbb E\big[\sum_{t=0}^{\infty}\gamma^t \mathbf 1\{x_t=\underline x\}\mid x_0,\pi_B\big]}\geq 0,
\end{equation*}
where $\underline x\in\arg\min_{x\in\mathcal X}\mathbb E\left[\sum_{t=0}^{\infty}\gamma^t\mathbf 1\{x_t=x\}\mid x_0,\pi_B\right]$.
They also show that by further restricting $\widetilde\epsilon(x)$ to be a constant function, the maximizer is given by 
\[
\widetilde \epsilon(x)=(1-\gamma)\cdot(d_0 - \mathcal D_{\pi_B}(x_0)), \,\,\forall x\in\mathcal X.
\]
Using the construction of the Lyapunov function $L_{\widetilde\epsilon}$,~\citet{chow2018lyapunov} propose the safe policy iteration (SPI) algorithm (see Algorithm~\ref{alg:safe_PI}) in which the Lyapunov function is updated via \emph{bootstrapping}, i.e.,~at each iteration $L_{\widetilde \epsilon}$ is recomputed using~\Eqref{eq:opt_eps_baseline} w.r.t.~the current baseline policy. At each iteration $k$, this algorithm has the following properties: {\bf 1)} {\em Consistent Feasibility}, i.e.,~if the current policy $\pi_k$ is feasible, then $\pi_{k+1}$ is also feasible; {\bf 2)} {\em Monotonic Policy Improvement}, i.e.,~$\mathcal C_{\pi_{k+1}}(x)\leq \mathcal C_{\pi_k}(x)$ for any $x\in\mathcal X$; and {\bf 3)} {\em Asymptotic Convergence}. Despite all these nice properties, SPI is still a value-function-based algorithm, and thus it is not straightforward to use it in continuous action problems. The main reason is that the greedification step becomes an optimization problem over the continuous set of actions that is not necessarily easy to solve. In Section~\ref{sec:algorthms}, we show how we use SPI and its nice properties to develop safe policy optimization algorithms that can handle continuous action problems. Our algorithms can be thought as combinations of DDPG or PPO (or any other on-policy or off-policy policy optimization algorithm) with a SPI-inspired critic that evaluates the policy and computes its corresponding Lyapunov function. The computed Lyapunov function is then used to guarantee safe policy update, i.e.,~the new policy is selected from a restricted set of safe policies defined by the Lyapunov function of the current policy. 

\begin{algorithm}[h!]
\begin{small}
\begin{algorithmic}
\STATE {\bf Input:} Initial feasible policy $\pi_0$;
\FOR{$k= 0,1,2,\ldots$}
\STATE {\bf Step 0:} With $\pi_b=\pi_k$, evaluate the Lyapunov function $L_{\epsilon_k}$, where $\epsilon_k$ is a solution of  \Eqref{eq:opt_eps_baseline}
\STATE {\bf Step 1:} Evaluate the cost value function $V_{\pi_k}(x)=\mathcal C_{\pi_k}(x)$; Then update the policy by solving the following problem:
$
\pi_{k+1}(\cdot|x)\in\arg\!\min_{\pi\in\mathcal F_{L_{\epsilon_k}}(x)}T_{\pi,c}[V_{\pi_k}](x),\forall x\in\mathcal X
$
\ENDFOR  
\STATE {\bf Return} Final policy $\pi_{k^*}$
\end{algorithmic}
\end{small}
\caption{Safe Policy Iteration (SPI)}
\label{alg:safe_PI}
\end{algorithm}

\newpage
\vspace{-0.05in}
\section{Lagrangian Approach to Safe RL}\label{appendix:lag}
\vspace{-0.05in}

There are a number of mild technical and notational assumptions which we will make throughout this section, so we state them here.
\begin{assumption}[Differentiability]\label{ass:differentiability}
For any state-action pair $(x,a)$, $\pi_\theta(a|x)$ is continuously
differentiable in $\theta$ and $\nabla_\theta\pi_\theta(a|x)$ is a
Lipschitz function in $\theta$ for every $a\in\mathcal A$ and $x\in\mathcal X$.
\end{assumption}
\begin{assumption}[Strict Feasibility]\label{ass:feasibility}
There exists a transient policy $\pi_\theta(\cdot|x)$ such that $\mathcal D_{\pi_\theta}(x_0)< d_0$ in the constrained problem.
\end{assumption}
\begin{assumption}[Step Sizes]\label{ass:steps}
The step size schedules $\{\alpha_{3,k}\}$, $\{\alpha_{2,k}\}$, and $\{\alpha_{1,k}\}$ satisfy
\begin{align}
\label{eq:step1_incre}
&\sum_k \alpha_{1,k} = \sum_k \alpha_{2,k} = \sum_k \alpha_{3,k} =\infty, \\
\label{eq:step2_incre}
&\sum_k \alpha_{1,k}^2,\;\;\;\sum_k \alpha_{2,k}^2,\;\;\;\sum_k \alpha_{3,k}^2<\infty, \\
\label{eq:step3_incre}
&\alpha_{1,k} = o\big(\alpha_{2,k}\big), \;\;\; \zeta_2(i) = o\big(\alpha_{3,k}\big).
\end{align}
\end{assumption}

Assumption~\ref{ass:differentiability} imposes smoothness on
the optimal policy. Assumption~\ref{ass:feasibility} guarantees the existence of
a local saddle point in the Lagrangian analysis introduced in the next
subsection. Assumption~\ref{ass:steps} refers to step sizes
corresponding to policy updates that will
be introduced for the algorithms in this paper, and indicates that the update
corresponding to $\{\alpha_{3,k}\}$ is on the fastest time-scale, the
updates corresponding to $\{\alpha_{2,k}\}$ is on the
intermediate time-scale, and the update corresponding to $\{\alpha_{1,k}\}$ is on
the slowest time-scale.  As this assumption refer to user-defined
parameters, they can always be chosen to be satisfied.

To solve the CMDP, we employ the Lagrangian relaxation procedure~\citep{bertsekas1999nonlinear} to convert it to the following unconstrained problem:  
\begin{equation}
\label{eq:unconstrained-discounted-risk-measure}
\max_{\lambda\geq 0}\min_{\theta}\bigg(L(\theta,\lambda)\stackrel{\triangle}{=} \mathcal C_{\pi_\theta}(x_0)+\lambda\left(\mathcal D_{\pi_\theta}(x_0)-d_0\right)\bigg),
\end{equation}
where $\lambda$ is the Lagrange multiplier.
Notice that $L(\theta,\lambda)$ is a linear function in $\lambda$. Then there exists a local saddle point $(\theta^*,\lambda^*)$  for the minimax optimization problem $\max_{\lambda\geq 0}\min_{\theta}L(\theta,\lambda)$, such that for some $r>0$, $\forall\theta\in\mathbb R^\kappa\cap B_{\theta^*}(r)$ and $\forall \lambda\in [0,\lambda_{\max}]$, we have
\begin{equation}\label{eq:local_saddle_point}
L(\theta,\lambda^*) \ge L( \theta^*,\lambda^*) \ge L( \theta^*,\lambda),
\end{equation}
where $B_{\theta^*}(r)$ is a hyper-dimensional ball centered at $\theta^*$ with radius $r>0$.

In the following, we present a policy gradient (PG) algorithm and an actor-critic (AC) algorithm. While the PG algorithm updates its parameters after observing several trajectories, the AC algorithms are incremental and update their parameters at each time-step.

We now present a policy gradient algorithm to solve the optimization problem~\Eqref{eq:unconstrained-discounted-risk-measure}. The idea of the algorithm is to descend in $\theta$ and ascend in $\lambda$ using the gradients of $L(\theta, \lambda)$ w.r.t.~$\theta$ and $\lambda$, i.e.,
\begin{align}
\label{eq:grad-theta}
\nabla_\theta L(\theta,\lambda) = \nabla_\theta \left(\mathcal C_{\pi_\theta}(x_0)+\lambda\mathcal D_{\pi_\theta}(x_0)\right), \,\nabla_\lambda L(\theta, \lambda) = \mathcal D_{\pi_\theta}(x_0)-d_0.
\end{align}

The unit of observation in this algorithm is a system trajectory generated by following policy $\pi_{\theta_k}$. At each iteration, the algorithm generates $N$ trajectories by following the current policy, uses them to estimate the gradients in \Eqref{eq:grad-theta}, and then uses these estimates to update the parameters $\theta,\lambda$.

Let $\xi=\{x_0,a_0,c_0,x_1,a_1,c_1,\ldots,x_{T-1},a_{T-1},c_{T-1},x_T\}$ be a trajectory generated by following the policy $\theta$, where $x_T=x_{\text{Tar}}$ is the target state of the system and $T$ is the (random) stopping time. The cost, constraint cost, and probability of $\xi$ are defined as $\mathcal C(\xi)=\sum_{k=0}^{T-1}\gamma^k c(x_k,a_k)$, $\mathcal D(\xi)=\sum_{k=0}^{T-1}\gamma^k d(x_k)$, and $\mathbb{P}_{\theta}(\xi)=P_0(x_0)\prod_{k=0}^{T-1}\pi_\theta(a_k|x_k)P(x_{k+1}|x_k,a_k)$, respectively. Based on the definition of $\mathbb{P}_{\theta}(\xi)$, one obtains $\nabla_\theta\log\mathbb{P}_{\theta}(\xi)=\sum_{k=0}^{T-1}\nabla_\theta\log\pi_\theta(a_k|x_k)$.

Algorithm~\ref{alg_traj} contains the pseudo-code of our proposed
policy gradient algorithm. What appears inside the parentheses on the
right-hand-side of the update equations are the estimates of the
gradients of $L(\theta,\lambda)$ w.r.t.~$\theta,\lambda$
(estimates of
the expressions in~\ref{eq:grad-theta}). Gradient
estimates of the Lagrangian function are given by
\[
\nabla_\theta L(\theta,\lambda)=\sum_{\xi} \mathbb{P}_\theta(\xi)\cdot\nabla_\theta\log\mathbb{P}_\theta(\xi)\left(\mathcal C_{\pi_\theta}(\xi) +\lambda\mathcal D_{\pi_\theta}(\xi)\right),\,\nabla_\lambda L(\theta,\lambda) = - d_0 +\sum_{\xi}\mathbb{P}_\theta(\xi)\cdot\mathcal D(\xi),
\]
where the likelihood gradient is
\begin{equation*}
\begin{split}
\nabla_\theta\log\mathbb{P}_\theta(\xi)=&\nabla_\theta\left\{\sum_{k=0}^{T-1} \log P(x_{k+1}|x_k,a_k)+\log\pi_\theta(a_k|x_k)+\log \mathbf 1\{x_0=x^0\}\right\}\\
=&\sum_{k=0}^{T-1}\nabla_\theta\log\pi_\theta(a_k|x_k)= \sum_{k=0}^{T-1}\frac{1}{\pi_\theta(a_k|x_k)}\nabla_\theta\pi_\theta(a_k|x_k).
\end{split}
\end{equation*}

In the algorithm, $\Gamma_\Lambda$ is a projection operator to
$[0,\lambda_{\max}]$, i.e.,
$\Gamma_\Lambda(\lambda)=\arg\min_{\hat\lambda\in[0,\lambda_{\max}]}\|\lambda-\hat\lambda\|^2_2$, which ensures the convergence of the
algorithm. Recall from Assumption~\ref{ass:steps} that the step-size schedules satisfy the standard conditions for stochastic approximation algorithms, and ensure that the policy parameter $\theta$ update is on the fast time-scale $\big\{\zeta_{2,i}\big\}$, and the Lagrange multiplier $\lambda$ update is on the slow time-scale $\big\{\zeta_{1,i}\big\}$. This results in a two time-scale stochastic approximation algorithm, which has shown to converge to a (local) saddle point of the objective function $L(\theta,\lambda)$. 
This convergence proof makes use of standard in many stochastic approximation theory, because in the limit when the step-size is sufficiently small, analyzing the convergence of PG is equivalent to analyzing the stability of an ordinary differential equation (ODE) w.r.t. its equilibrium point.

In policy gradient, the unit of observation is a system
trajectory. This may result in high variance for the gradient
estimates, especially when the length of the trajectories is long. To
address this issue,  we propose two actor-critic
algorithms that use value function approximation in the
gradient estimates and update the parameters incrementally (after each
state-action transition). We present two actor-critic algorithms for
optimizing ~\Eqref{eq:unconstrained-discounted-risk-measure}. These
algorithms are still based on the above gradient estimates. Algorithm~\ref{alg:AC} contains the pseudo-code of these
algorithms. The projection operator
$\Gamma_\Lambda$ is
necessary to ensure the convergence of the algorithms. Recall from
Assumption~\ref{ass:steps} that the step-size schedules satisfy the standard conditions for stochastic approximation algorithms, and ensure that the critic update is on the fastest time-scale $\big\{\alpha_{3,k}\big\}$, the policy and $\theta$-update $\big\{\alpha_{2,k}\big\}$ is on the intermediate timescale, and finally the Lagrange multiplier update is on the slowest time-scale $\big\{\alpha_{1,k}\big\}$. This results in three time-scale stochastic approximation algorithms. 

Using the policy gradient theorem from \citet{Sutton00PG}, one can show that
\begin{equation}
\label{eq:PG-Thm}
\nabla_\theta L(\theta,\lambda) = \nabla_\theta V_\theta(x_0) = \frac{1}{1-\gamma}\sum_{x,a}\mu_{\theta}(x,a|x_0)\;\nabla\log\pi_\theta(a|x)\;Q_\theta(x,a), 
\end{equation}
where $\mu_{\theta}$ is the discounted visiting distribution and $Q_\theta$ is the action-value function of policy $\theta$. We can show that $\frac{1}{1-\gamma}\nabla\log\pi_\theta(a_k|x_k)\cdot\delta_k$ is an unbiased estimate of $\nabla_\theta L(\theta,\lambda)$, where 
\[
\delta_k=c_{\lambda}(x_k,a_k)+\gamma\widehat{V}_\theta(x_{k+1})-\widehat{V}_\theta(x_k)
\]
is the temporal-difference (TD) error, and $\widehat{V}_\theta$ is the value estimator of $V_\theta$. 

Traditionally, for convergence guarantees in actor-critic algorithms, the critic uses linear approximation for the value function $V_\theta(x)\approx v^\top\psi(x)=\widehat{V}_{\theta,v}(x)$, where the feature vector $\psi(\cdot)$ belongs to a low-dimensional space $\mathbb R^{\kappa_2}$. The linear approximation $\widehat{V}_{\theta,v}$ belongs to a low-dimensional subspace
$S_{V}=\left\{\Psi v|v\in\mathbb R^{\kappa_2}\right\}$, where $\Psi$ is a short-hand notation for the set of features, i.e., $\Psi(x)=\psi^\top(x)$. Recently with the advances of deep neural networks, it has become increasingly popular to model the critic with a deep neural network architecture, based on the objective function of minimizing the MSE of Bellman residual w.r.t. $V_\theta$ or $Q_\theta$ \citep{mnih2013playing}.  

\newpage
\vspace{-0.05in}
\section{Experimental Setup in MuJoCo Tasks}\label{appendix:exp}
\vspace{-0.05in}
Our experiments are performed on safety-augmented versions of standard MuJoCo domains~\citep{mujoco}.

{\bf HalfCheetah-Safe.} The agent is a the standard HalfCheetah (a 2-legged simulated robot rewarded for running at high speed) augmented with safety constraints.  We choose the safety constraints to be defined on the speed limit.  We constrain the speed to be less than $1$, i.e., constraint cost is thus $\mathbf 1[|v|>1]$ . Episodes are of length $200$. The constraint threshold is $50$.

{\bf Point Circle.} This environment is taken from~\citep{achiam2017constrained}. The agent is a point mass (controlled via a pivot).  The agent is initialized at $(0,0)$ and rewarded for moving counter-clockwise along a circle of radius $15$ according to the reward $\frac{-dx \cdot y + dy\cdot x}{1 + |\sqrt{x^2 + y^2} - 15|}$, for position $x,y$ and velocity $dx,dy$.  The safety constraint is defined as the agent staying in a position satisfying $|x|\le 2.5$.  The constraint cost is thus $\mathbf 1[|x|>2.5]$.  Episodes are of length $65$. The constraint threshold is $7$.

{\bf Point Gather.} This environment is taken from~\citep{achiam2017constrained}. The agent is a point mass (controlled via a pivot) and the environment includes randomly positioned apples (2 apples) and bombs (8 bombs).  The agent given a reward of $10$ for each apple collected and a penalty of $-10$ for each bomb.  The safety constraint is defined as number of bombs collected during the episode. Episodes are of length $15$. The constraint threshold is $4$ for DDPG and $2$ for PPO.


{\bf Ant Gather.} This environment is is the same as Point Circle, only with an ant agent (quadrapedal simulated robot).  Each episode is initialized with 8 apples and 8 bombs.  The agent given a reward of $10$ for each apple collected, a penalty reward of $-20$ for each bomb collected, and a penalty reward of $-20$ is incurred if the episode terminates prematurely (because the ant falls). Episodes are of length at most $500$. The constraint threshold is $10$ for DDPG algorithms and is $5$ for PPO algorithms.

\begin{figure}[h!]
\begin{center}
  \begin{tabular}{ccccc}
    \small HalfCheetah-Safe & \small Point-Circle & \small Ant-Gather & \small Point-Gather  \\
    \includegraphics[width=0.2\columnwidth]{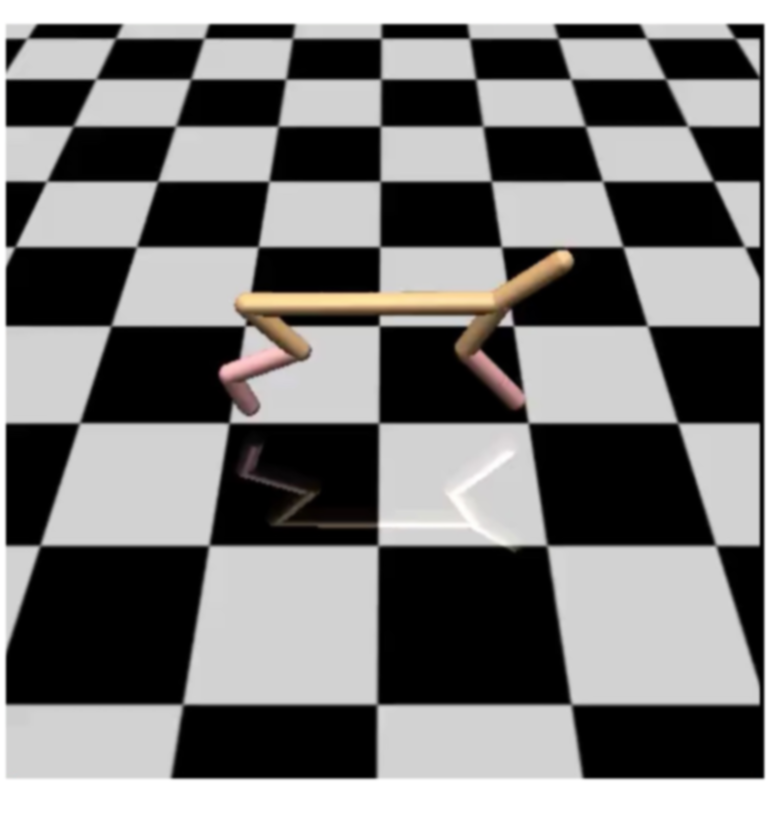} &
    \includegraphics[width=0.23\columnwidth]{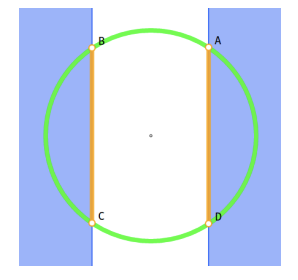} &
    \includegraphics[width=0.26\columnwidth]{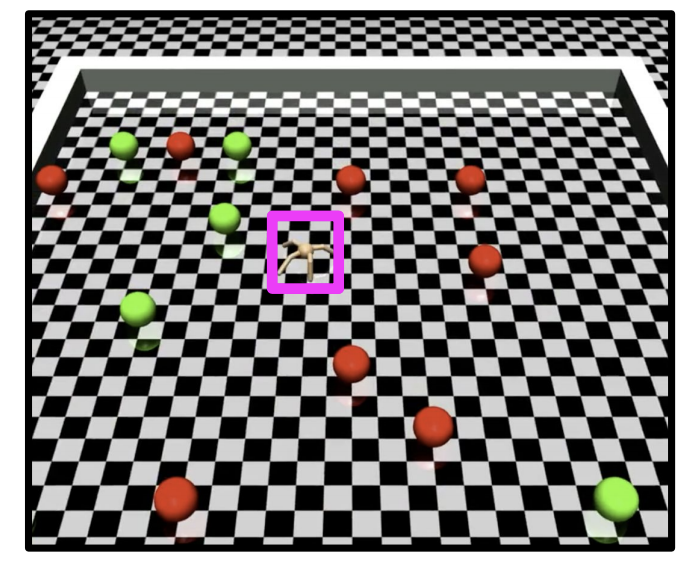}  &
    \includegraphics[width=0.22\columnwidth]{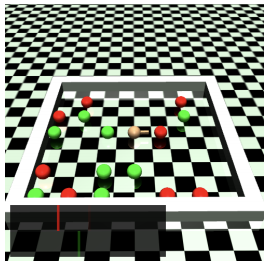} 
  \end{tabular}
\end{center}
\caption{
The Robot Locomotion Control Tasks
}
\label{fig:robot_task}
\end{figure}

Figure \ref{fig:robot_task} shows the visualization of the above domains used in our experiments.

In these experiments there are three different agents: (1) a point-mass
($\mathcal X\subseteq \mathbb R^9$, $A\subseteq\mathbb R^2$); an ant quadruped robot ($\mathcal X\subseteq \mathbb R^{32}$, $A\subseteq\mathbb R^8$); (3) a half-cheetah ($\mathcal X\subseteq \mathbb R^{18}$, $A\subseteq\mathbb R^6$). 
For all experiments,  we use two neural networks with
two hidden layers of size $(100, 50)$ and ReLU activation to model the mean and log-variance of the Gaussian actor policy, and two neural networks with
two hidden layers of size $(200, 50)$ and $\tanh$ activation to model the critic and constraint critic. To build a low variance sample gradient estimate, we use $\text{GAE-}\lambda$ \citep{schulman2015high} to estimate the advantage and constraint advantage functions, with a hyper-parameter $\lambda\in(0,1)$ optimized by grid-search.

On top of the GAE parameter $\lambda$, in all numerical experiments and for each algorithm (SPPO $\theta$-projection, SDDPG $\theta$-projection, SPPO $a$-projection, SDDPG $a$-projection, CPO, Lagrangian, and the unconstrained PG counterparts), we systematically explored different parameter settings by doing grid-search over the following factors: (i) learning rates in the actor-critic algorithm, (ii) batch size, (iii) regularization parameters of the policy relative entropy term, (iv) with-or-without natural policy gradient updates, (v) with-or-without the emergency safeguard PG updates (see Appendix \ref{sec:practical} for more details). Although each algorithm might have a different parameter setting that leads to the optimal performance in training, the results reported here are the best ones for each algorithm, chosen by the same criteria (which is based on value of return plus certain degree of constraint satisfaction). To account for the variability during training, in each learning curve a $1$-SD confidence interval is also computed over $10$ separate random runs (under the same parameter setting).

\subsection{More Explanations on MuJoCo Results}
In all numerical experiments and for each algorithm (SPPO $\theta$-projection, SDDPG $\theta$-projection, SPPO $a$-projection, SDDPG $a$-projection, CPO, Lagrangian, and the unconstrained PG counterparts), we systematically explored various hyper-parameter settings by doing grid-search over the following factors: (i) learning rates in the actor-critic algorithm, (ii) batch size, (iii) regularization parameters of the policy relative entropy term, (iv) with-or-without natural policy gradient updates, (v) with-or-without the emergency safeguard PG updates (see Appendix \ref{sec:practical} for more details). Although each algorithm might have a different parameter setting that leads to the optimal training performance, the results reported in the paper are the best ones for each algorithm, chosen by the same criteria (which is based on value of return + certain degree of constraint satisfaction). 

In our experiments, we compare the two classes of safe RL algorithms, one derived from $\theta$-projection (constrained policy optimization) and one from the $a$-projection (safety layer), with the unconstrained and Lagrangian baselines in four problems: PointGather, AntGather, PointCircle, and HalfCheetahSafe. We perform these experiments with both off-policy (DDPG) and on-policy (PPO) versions of the algorithms. 

In PointCircle DDPG, although the Lagrangian algorithm significantly outperforms the safe RL algorithms in terms of return, it violates the constraint more often. The only experiment in which Lagrangian performs similarly to the safe algorithms in terms of both return and constraint violation is PointCircle PPO. 
In all other experiments that are performed in the HalfCheetahSafe, PointGather and AntGather domains, either (i) the policy learned by Lagrangian has a significantly lower performance than that learned by one of the safe algorithms (see HalfCheetahSafe DDPG, PointGather DDPG, AntGather DDPG), or (ii) the Lagrangian method violates the constraint during training, while the safe algorithms do not (see HalfCheetahSafe PPO, PointGather PPO, AntGather PPO). This clearly illustrates the effectiveness of our Lyapunov-based safe RL algorithms, when compared to Lagrangian method.


\newpage
\vspace{-0.05in}
\section{Experimental Setup in Robot Navigation}\label{appendix:robot}
\vspace{-0.05in}
P2P is a continuous control task with a goal of navigating a robot to any arbitrary goal position collision-free and without memory of the workspace topology. The goal is usually within $5-10$ meters from the robot agent, but it is not visible to the agent before the task starts, due to both limited sensor range and the presence of obstacles that block a clear line of sight. The agent's observations, $\vx = (\vg, \dot{\vg}, \vl) \in\mathbb R^{68},$ consists of the relative goal position, the relative goal velocity, and the Lidar measurements. Relative goal position, $\vg,$ is the relative polar coordinates between the goal position and the current robot pose, and $\dot{\vg}$ is the time derivative of $\vg$, which indicates the speed of the robot navigating to the goal. This information is available from the robot's localization sensors. Vector $\vl$ is the noisy Lidar input (Fig. \ref{fig:lidar}), which measures the nearest obstacle in a direction within a $220^{\circ}$ field of view split in $64$ bins, up to $5$ meters in depth. The action is given by $\va \in \mathbb R^2$, which is linear and angular velocity vector at the robot's center of the mass. The transition probability $P:\mathcal X\times\mathcal A\rightarrow\mathcal X$ captures the noisy differential drive robot dynamics. Without knowing the full non-linear system dynamics, we here assume knowledge of a simplified blackbox kinematics simulator operating at $5$Hz in which Gaussian noise, $\mathcal N(0, 0.1),$ is added to both the observations and actions in order to model the noise in sensing, dynamics, and action actuations in real-world. The objective of the P2P task is to navigate the robot to reach within $30$ centimeters from any real-time goal. 
While the dynamics of this system is simpler than that of HalfCheetah. But unlike the MuJoCo tasks where the underlying dynamics are deterministic, in this robot experiment the sensor, localization, and dynamics noise paired with partial world observations and unexpected obstacles make this safe RL much more challenging. More descriptions about the indoor robot navigation problem and its implementation details can be found in Section 3 and 4 of \citet{autorl}.

Here the CMDP is non-discounting and has a finite-horizon of $T=100$. We reward the agent for reaching the goal, which translates to an immediate cost of $c(\vx,\va)=\|g\|^2$, which measures the relative distance to goal.
To measure the impact energy of obstacle collisions, we impose an immediate constraint cost of $d(\vx,\va)=\|\dot{\vg}\| \cdot \mathbf 1\{\|\vl\|\leq r_{\text{impact}}\} /T$, where $r_{\text{impact}}$ is the impact radius w.r.t. the Lidar depth signal, to account for the speed during collision, with a constraint threshold $d_0$ that characterizes the agent's maximum tolerable collision impact energy to any objects. (Here the total impact energy is proportional to the robot's speed during any collisions.) 
Under this CMDP framework (Fig. \ref{fig:sddpg}), the main goal is to train a policy $\pi^*$ that drives the robot along the shortest path to the goal and to limit the average impact energy of obstacle collisions. Furthermore, due to limited data any intermediate point-to-point policy is deployed on the robot to collect more samples for further training, therefore guaranteeing safety during training is critical in this application.

\begin{figure*}[h]
\centering
\begin{tabular}{ccc}
    \subfloat[\scriptsize Training, 23 by 18m]{\includegraphics[trim=0mm 0mm 0mm 0mm,clip,width=0.45\textwidth,keepaspectratio=true]{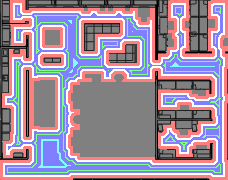}\label{fig:trainingEnv}}&
    \subfloat[\scriptsize Building 2, 60 by 47m]{\includegraphics[trim=0mm 0mm 0mm 0mm,clip,width=0.45\textwidth,keepaspectratio=true]{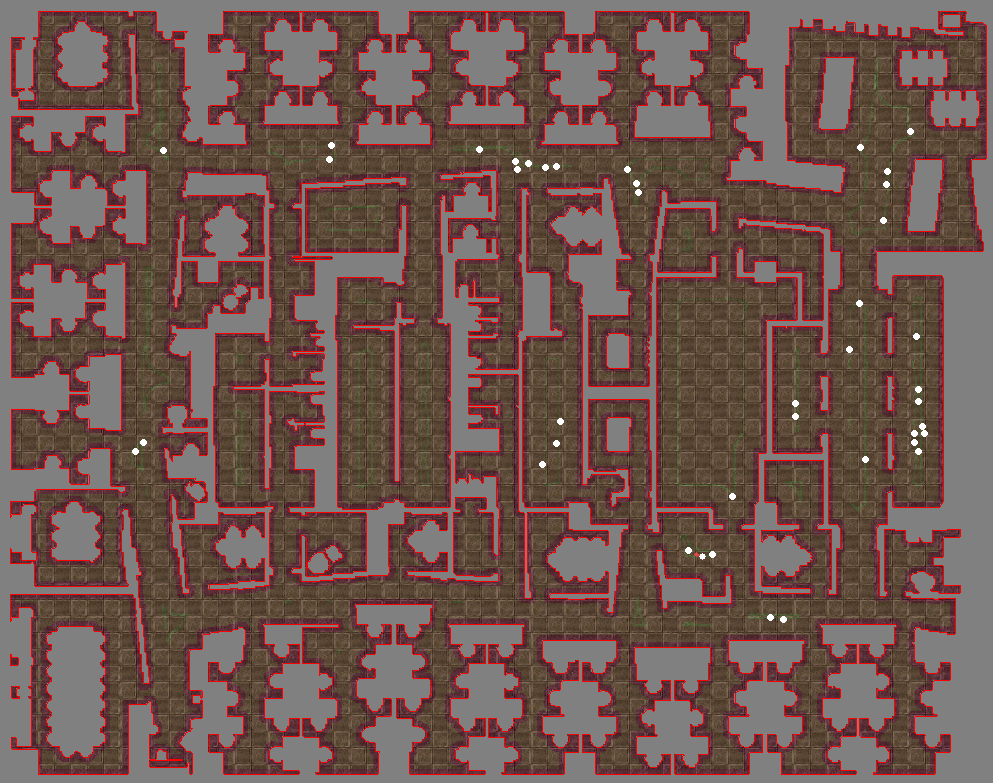}\label{fig:sfo1ms}}&
\end{tabular}
    \caption{\footnotesize (a) Training and (b) evaluation environments, generated from real office building plans. The evaluation environment is an order of magnitude bigger. \label{fig:environments}}
\end{figure*}

\begin{figure*}[tb]
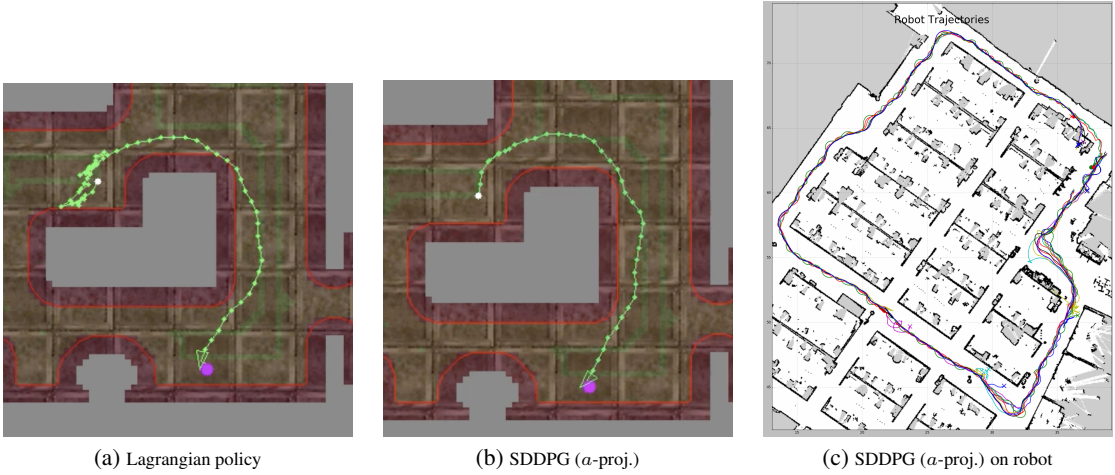

  \begin{tabular}{ccc}
    \subfloat[\scriptsize Lagrangian policy]{\label{fig:lag-nav1}\includegraphics[width=0.27\textwidth]{lagddpg_route.png}}\quad &
    \subfloat[\scriptsize SDDPG ($a$-proj.)]{\includegraphics[width=0.27\textwidth]{sddpg_route.png}\label{fig:lyap-nav1}}\quad &
    \subfloat[\scriptsize SDDPG ($a$-proj.) on robot]{\includegraphics[width=0.27\textwidth]{robot_traj2.png}\label{fig:on-robot1}}    
  \end{tabular}
      \vspace{-0.1in}
   \caption{\footnotesize 
Navigation routes of two policies on a similar setup (a) and (b). Log of on-robot experiments (c).
\label{fig:robot_ddpg1}}
\end{figure*}

\newpage
\vspace{-0.05in}
\section{Pseudo-code of the Safe Policy Gradient Algorithms}\label{appendix:alg}
\vspace{-0.05in}
 \begin{algorithm}[h!]
\begin{algorithmic}
\STATE {\bf Input:} parameterized policy $\pi(\cdot|\cdot;\theta)$
\STATE {\bf Initialization:} policy parameter $\theta=\theta_0$, and the Lagrangian parameter $\lambda=\lambda_0$
\FOR{$i = 0,1,2,\ldots$}
\FOR{$j = 1,2,\ldots$}
\STATE Generate $N$ trajectories $\{\xi_{j,i}\}_{j=1}^N$ by starting at $x_0$ and following the policy $\theta_i$.
\ENDFOR  
\vspace{-0.2in}
\begin{small}
\begin{align*}
\textrm{\bf $\theta$ Update:} \quad &  \theta_{i+1} = \theta_i -\alpha_{2,i}\frac{1}{N}\sum_{j=1}^N\nabla_\theta\log\mathbb{P}_\theta(\xi_{j,i})\vert_{\theta=\theta_i}\left(\mathcal C(\xi_{j,i}) +\lambda_i\mathcal D(\xi_{j,i})\right) \nonumber \\ 
\textrm{{\bf $\lambda$ Update:}}\quad&\lambda_{i+1} = \Gamma_\Lambda\bigg[\lambda_i + \alpha_{1,i}\bigg( - d_0 + \frac{1}{N}\sum_{j=1}^N\mathcal D(\xi_{j,i})\bigg)\bigg]
\end{align*}
\end{small}
\vspace{-0.2in}
\ENDFOR  
\end{algorithmic}
\caption{Lagrangian Trajectory-based Policy Gradient Algorithm}
\label{alg_traj}
\end{algorithm}

\begin{algorithm}
\begin{small}
\begin{algorithmic}[h!]
\STATE {\bf Input:} Parameterized policy $\pi(\cdot|\cdot;\theta)$ and value function feature vector $\phi(\cdot)$
\STATE {\bf Initialization:} policy parameters $\theta=\theta_0$; Lagrangian parameter $\lambda=\lambda_0$; value function weight $v=v_0$
\WHILE{TRUE}
\FOR{$k = 0,1,2,\ldots$}
\STATE Sample $\;a_k\sim\pi(\cdot|x_k;\theta_k)$; $\;c_{\lambda_k}(x_k,a_k)=c(x_k,a_k)+\lambda_k d(x_k)$; $\,x_{k+1}\sim {P}(\cdot|x_k,a_k)$; $\;$
\STATE \textbf{// AC Algorithm:}
\vspace{-0.05in}
\begin{small}
\begin{align}
\textrm{\bf TD Error:} \quad & \delta_k(v_k) = c_{\lambda_k}(x_k,a_k) + \gamma \widehat V_{\phi_k}(x_{k+1}) - \widehat V_{\phi_k}(x_k) \label{TD-calc} \\
\textrm{\bf Critic Update:} \quad & v_{k+1}=v_k+\zeta_3(k)\delta_k(v_k)\psi(x_k) \label{v_up_incre} \\
\textrm{{\bf $\theta$ Update:}}\quad &\theta_{k+1} = \theta_k-{\zeta_2(k)}\nabla_\theta\log\pi_\theta(a_k|x_k)\cdot\delta_k(v_k)/{1-\gamma} \label{theta_up_incre} \\
\textrm{{\bf $\lambda$ Update:}}\quad &\lambda_{k+1} = \Gamma_\Lambda\Big(\lambda_k + \zeta_1(k)\big(-d_0 + \frac{1}{N}\sum_{j=1}^N\mathcal D(\xi_{j,i})\big)\Big) \label{lambda_up_incre}
\end{align}
\end{small}
\vspace{-0.2in}
\STATE \textbf{// NAC Algorithm:}
\vspace{-0.1in}
\begin{small}
\begin{align}
\textrm{{\bf Critic Update:}}\quad &w_{k+1} = \left(I-\zeta_3(k)\nabla_\theta \log\pi_\theta(a_k|x_k)\vert_{\theta=\theta_k}\left(\nabla_\theta \log\pi_\theta(a_k|x_k)\vert_{\theta=\theta_k}\right)^\top\right)w_k\nonumber\\
&\quad\quad\,\,+\zeta_3(k)\delta_k(v_k)\nabla_\theta \log\pi_\theta(a_k|x_k)\vert_{\theta=\theta_k} \label{w_NAC_up_incre} \\
\textrm{{\bf $\theta$ Update:}}\quad &\theta_{k+1} =  \theta_k-{\zeta_2(k)w_k}/{1-\gamma} \label{theta_NAC_up_incre} \\
\textrm{{\bf Other Updates:}}\quad & \textrm{Follow from Eqs.~\ref{TD-calc},~\ref{v_up_incre}, and~\ref{lambda_up_incre}.}\nonumber
\end{align}
\end{small}
\ENDFOR
\ENDWHILE
\end{algorithmic}
\end{small}
\caption{Lagrangian Actor-Critic Algorithm}
\label{alg:AC}
\end{algorithm}

\begin{algorithm}[h!]
\begin{small}
\begin{algorithmic}
\STATE {\bf Input:} Initial feasible policy $\pi_0$;
\FOR{$k= 0,1,2,\ldots$}
\STATE {\bf Step 0:} With $\pi_b=\pi_{\theta_k}$, generate $N$ trajectories $\{\xi_{j,k}\}_{j=1}^N$ of $T$ steps by starting at $x_0$ and following the policy $\theta_k$
\STATE {\bf Step 1:} Using the trajectories $\{\xi_{j,k}\}_{j=1}^N$, estimate the critic $Q_{\theta}(x,a)$ and the constraint critic $Q_{D,\theta}(x,a)$; 
\begin{itemize}
\item For DDPG, these functions are trained by minimizing the MSE of Bellman residual, and one can also use off-policy samples from replay buffer \citep{schaul2015prioritized}; 
\item For PPO these functions can be estimated by the generalized advantage function technique from \cite{schulman2015high}
\end{itemize}
\STATE {\bf Step 2:} Based on the closed form solution of a QP problem with an LP constraint in Section 10.2 of \cite{achiam2017constrained}, calculate $\lambda^*_k$ with the following formula:
\begin{small}
\[
\lambda^*_k=\left(\frac{-\beta_k\widetilde\epsilon-\left(\nabla_\theta Q_{\theta}(\bar x,\bar a)\mid_{\theta=\theta_k}\right)^\top H(\theta_k)^{-1}\nabla_\theta Q_{D,\theta}(\bar x,\bar a)\mid_{\theta=\theta_k}}{\left(\nabla_\theta Q_{D,\theta}(\bar x,\bar a)\mid_{\theta=\theta_k}\right)^\top H(\theta_k)^{-1}\nabla_\theta Q_{D,\theta}(\bar x,\bar a)\mid_{\theta=\theta_k}}\right)_+,
\]
\end{small}
where 
\begin{small}
\[
\begin{split}
&\nabla_\theta Q_{\theta}(\bar x,\bar a)=\frac{1}{N}\sum_{x,a\in\xi_{j,k},1\leq j\leq N}\sum_{t=0}^{T-1}\gamma^t\nabla_\theta \log\pi_\theta(a|x) Q_{\theta}(x, a),\\
&\nabla_\theta Q_{D,\theta}(\bar x,\bar a)=\frac{1}{N}\sum_{x,a\in\xi_{j,k},1\leq j\leq N}\sum_{t=0}^{T-1}\gamma^t\nabla_\theta \log\pi_\theta(a|x) Q_{\theta}( x, a),
\end{split}
\]
\end{small}
$\beta_k$ is the adaptive penalty weight of the $\overline D_{\text{KL}}(\pi||\pi_{\theta_k})$ regularizer, and $H(\theta_k)=\nabla_\theta^2 \overline D_{\text{KL}}(\pi||\pi_{\theta})\mid_{\theta=\theta_k}$ is the Hessian of this term
\STATE {\bf Step 3:} Update the policy parameter by following the objective gradient; 
\begin{itemize}
\item For DDPG 
\[
\theta_{k+1}\leftarrow\theta_k-\alpha_{k}\cdot\frac{1}{N \cdot T}\sum_{x\in\xi_{j,k},1\leq j\leq N}\nabla_\theta\pi_\theta(x)\mid_{\theta=\theta_k}\cdot(\nabla_a Q_{\theta_k}(x,a) + \lambda^*_k\nabla_a Q_{D,\theta_k}(x,a))\mid_{a=\pi_{\theta_k}(x)}
\]
\item For PPO, 
\[
\begin{split}
\theta_{k+1}\leftarrow\theta_k-\frac{\alpha_{k}}{N\beta_k}\left(H(\theta_k)\right)^{-1}\!\!\!\!\!\sum_{x_{j,t}, a_{j,t}\in\xi_{j,k},1\leq j\leq N}\sum_{t=0}^{T-1}&\gamma^t\cdot\nabla_\theta\log\pi_\theta(a_{j,t}|x_{j,t})\mid_{\theta=\theta_k}\cdot \\
&(Q_{\theta_k}(x_{j,t,}a_{j,t})+\lambda^*_kQ_{D,\theta_k}(x_{j,t,}a_{j,t}))
\end{split}
\]
\end{itemize}
\STATE {\bf Step 4:} At any given state $x\in\mathcal X$, compute the feasible action probability $a^*(x)$ via action projection in the safety layer, that takes inputs $\nabla_a Q_L(x,a)=\nabla_a Q_{D,\theta_k}(x,a)$ and $\epsilon(x)=(1-\gamma)(d_0-Q_{D,\theta_k}(x_0,\pi_k(x_0)))$, for any $a\in\mathcal A$.

\ENDFOR  
\STATE {\bf Return} Final policy $\pi_{\theta_{k^*}}$, \end{algorithmic}
\end{small}
\caption{Lyapunov-based Policy Gradient with $\theta$-projection  (SDDPG and SPPO)}
\label{alg:cpo}
\end{algorithm}

\begin{algorithm}[h!]
\begin{small}
\begin{algorithmic}
\STATE {\bf Input:} Initial feasible policy $\pi_0$;
\FOR{$k= 0,1,2,\ldots$}
\STATE {\bf Step 0:} With $\pi_b=\pi_{\theta_k}$, generate $N$ trajectories $\{\xi_{j,k}\}_{j=1}^N$ of $T$ steps by starting at $x_0$ and following the policy $\theta_k$
\STATE {\bf Step 1:} Using the trajectories $\{\xi_{j,k}\}_{j=1}^N$, estimate the critic $Q_{\theta}(x,a)$ and the constraint critic $Q_{D,\theta}(x,a)$; 
\begin{itemize}
\item For DDPG, these functions are trained by minimizing the MSE of Bellman residual, and one can also use off-policy samples from replay buffer \citep{schaul2015prioritized}; 
\item For PPO these functions can be estimated by the generalized advantage function technique from \cite{schulman2015high}
\end{itemize}
\STATE {\bf Step 2:} Update the policy parameter by following the objective gradient; 
\begin{itemize}
\item For DDPG 
\[
\theta_{k+1}\leftarrow\theta_k-\alpha_{k}\cdot\frac{1}{N \cdot T}\sum_{x\in\xi_{j,k},1\leq j\leq N}\nabla_\theta\pi_\theta(x)\mid_{\theta=\theta_k}\cdot\nabla_a Q_{\theta_k}(x,a)\mid_{a=\pi_{\theta_k}(x)};
\]
\item For PPO, 
\[
\theta_{k+1}\leftarrow\theta_k-\frac{\alpha_{k}}{N\beta_k}\left(H(\theta_k)\right)^{-1}\!\!\!\!\!\sum_{x_{j,t}, a_{j,t}\in\xi_{j,k},1\leq j\leq N}\sum_{t=0}^{T-1}\gamma^t\cdot\nabla_\theta\log\pi_\theta(a_{j,t}|x_{j,t})\mid_{\theta=\theta_k}\cdot Q_{\theta_k}(x_{j,t,}a_{j,t})
\]
where $\beta_k$ is the adaptive penalty weight of the $\overline D_{\text{KL}}(\pi||\pi_{\theta_k})$ regularizer, and $H(\theta_k)=\nabla_\theta^2 \overline D_{\text{KL}}(\pi||\pi_{\theta})\mid_{\theta=\theta_k}$ is the Hessian of this term
\end{itemize}
\STATE {\bf Step 3:} At any given state $x\in\mathcal X$, compute the feasible action probability $a^*(x)$ via action projection in the safety layer, that takes inputs $\nabla_a Q_L(x,a)=\nabla_a Q_{D,\theta_k}(x,a)$ and $\epsilon(x)=(1-\gamma)(d_0-Q_{D,\theta_k}(x_0,\pi_k(x_0)))$, for any $a\in\mathcal A$.

\ENDFOR  
\STATE {\bf Return} Final policy $\pi_{\theta_{k^*}}$, \end{algorithmic}
\end{small}
\caption{Lyapunov-based Policy Gradient with $a$-projection (SDDPG-modular and SPPO-modular)}
\label{alg:safety_layer}
\end{algorithm}

\subsection{Practical Implementations of Safe PG}\label{sec:practical}

Due to function approximation errors, even with the Lyapunov constraints in practice the safe PG algorithm may take a bad step and
produce an infeasible policy update and cannot automatically recover from such a bad
step. To tackle this issue, similar to \citet{achiam2017constrained} we propose the following \emph{safeguard} policy update rule to purely decrease
the constraint cost:
$
\theta_{k+1}=\theta_k-\alpha_{\text{sg},k}{\nabla}_\theta\mathcal D_{\pi_\theta}(x_0)_{\theta=\theta_k},
$
where $\alpha_{\text{sg},k}$ is the learning rate for safeguard update. If $\alpha_{\text{sg},k}>>\alpha_{k}$ (learning rate of PG), then with the safeguard update $\theta$ will quickly recover from the bad step but it might be overly conservative. 
This approach
is principled because as soon as $\pi_{\theta_k}$ is unsafe/infeasible w.r.t. CMDP, the algorithm uses a limiting search direction. One can directly extend this safeguard update to the multiple-constraint scenario by doing gradient descent over the constraint that has the worst violation. 
Another remedy to reduce the chance of constraint violation is to do \emph{constraint tightening} on the constraint cost threshold. Specifically, instead of $d_0$, one may pose the constraint based on $d_0\cdot (1-\delta)$, where $\delta\in(0,1)$ is the factor of safety for providing additional buffer to constraint violation.
Additional techniques in cost-shaping have been proposed in \citet{achiam2017constrained} to smooth out the sparse constraint costs. While these techniques can further ensure safety, construction of the cost-shaping term requires knowledge from the environment, which makes the safe PG algorithms more complicated.

\end{document}